%% file: main.tex
\documentclass[]{holiday}

\title{FlashSAC: Fast and Stable Off-Policy Reinforcement Learning for High-Dimensional Robot Control}

\author[1,*]{Donghu Kim}
\author[2,3,*]{Youngdo Lee}
\author[2]{Minho Park}
\author[2]{Kinam Kim}
\author[3]{I Made Aswin Nahendra}
\author[4]{Takuma Seno}
\author[1]{Sehee Min}
\author[5,6]{Daniel Palenicek}
\author[7]{Florian Vogt}
\author[7]{Danica Kragic}
\author[5,6,8,9]{Jan Peters}
\author[2,\dagger]{Jaegul Choo}
\author[1,\dagger]{Hojoon Lee}

\affiliation[1]{Holiday Robotics}
\affiliation[2]{KAIST}
\affiliation[3]{KRAFTON}
\affiliation[4]{Turing Inc}
\affiliation[5]{TU Darmstadt}
\affiliation[6]{hessian.AI}
\affiliation[7]{KTH Royal Institute of Technology}
\affiliation[8]{German Research Center for AI (DFKI)} 
\affiliation[9]{Robotics Institute Germany (RIG)}

\contribution[*]{Equal Contribution}
\contribution[\dagger]{Joint Correspondence}

\abstract{\subsubsection*{Abstract} 
Reinforcement learning (RL) is a core approach for robot control when expert demonstrations are unavailable. 
On-policy methods such as Proximal Policy Optimization (PPO) are widely used for their stability, but their reliance on narrowly distributed on-policy data limits accurate policy evaluation in high-dimensional state and action spaces. 
Off-policy methods can overcome this limitation by learning from a broader state-action distribution, yet suffer from slow convergence and instability, as fitting a value function over diverse data requires many gradient updates, causing critic errors to accumulate through bootstrapping.
We present \textsc{FlashSAC}, a fast and stable off-policy RL algorithm built on Soft Actor-Critic. Motivated by scaling laws observed in supervised learning, 
\textsc{FlashSAC} sharply reduces gradient updates while compensating with larger models and higher data throughput. 
To maintain stability at increased scale, \textsc{FlashSAC} explicitly bounds weight, feature, and gradient norms, curbing critic error accumulation.
Across over 60 tasks in 10 simulators, \textsc{FlashSAC} consistently outperforms PPO and strong off-policy baselines in both final performance and training efficiency, with the largest gains on high-dimensional tasks such as dexterous manipulation. In sim-to-real humanoid locomotion, \textsc{FlashSAC} reduces training time from hours to minutes, demonstrating the promise of off-policy RL for sim-to-real transfer.
}

\correspondence{\email{\{donghu.kim, hojoon.lee\}@holiday-robotics.com}}

\metadata[Webpage]{\url{https://holiday-robot.github.io/FlashSAC}}
\date{\today}

\begin{document}

\maketitle

\vspace{5.5mm}
\begin{center}
    \captionsetup{type=figure}
    \includegraphics[width=0.999\textwidth]{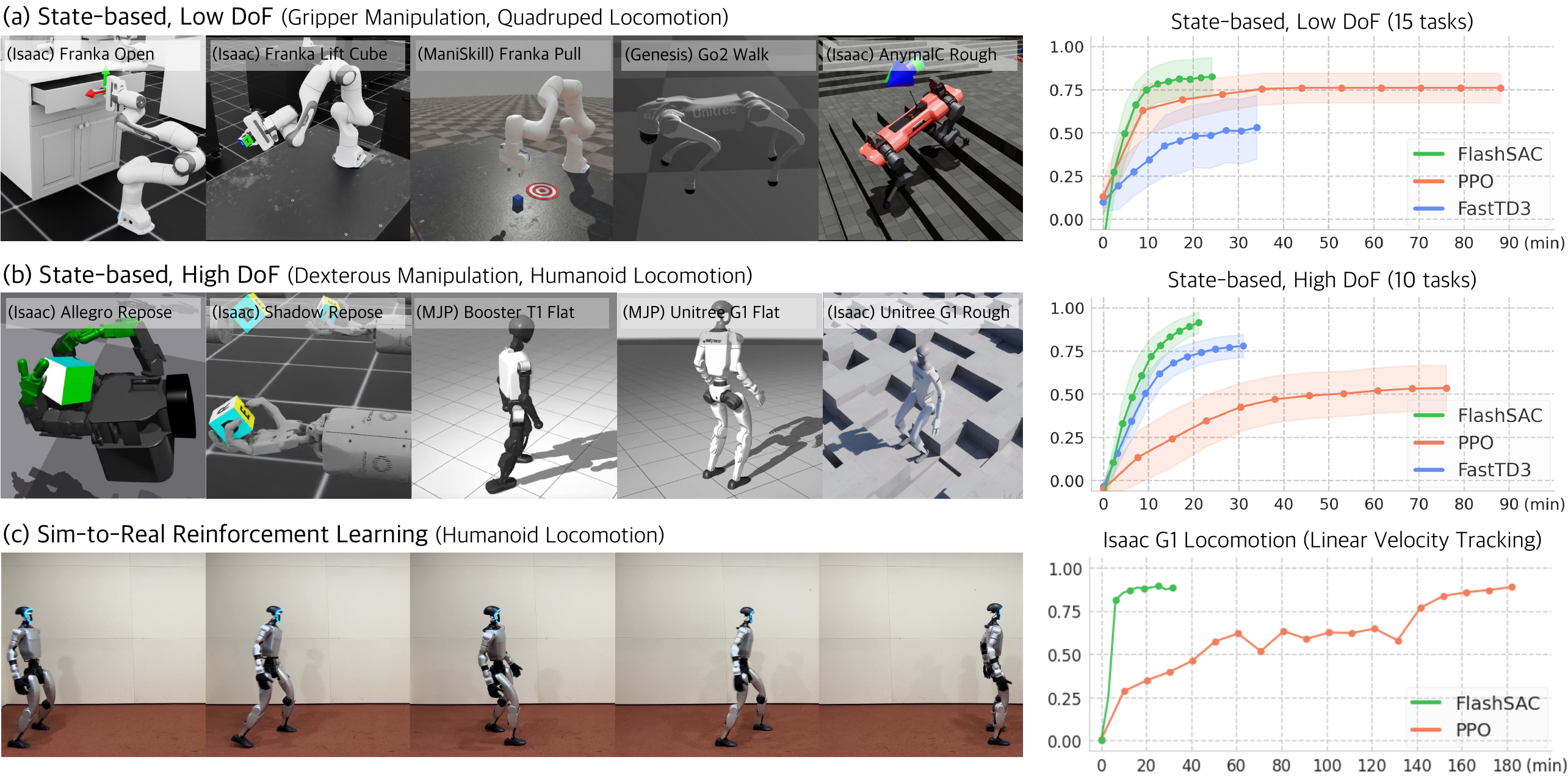}
    \vspace{-6mm}
    \caption{
        \textbfs{Results Overview.}
        Tasks grouped by state–action dimensionality, with representative examples shown for each category.
        \textbfs{(a) State-based, Low DoF:} Gripper manipulation and quadruped locomotion tasks from IsaacLab, ManiSkill, and Genesis. In low-dimensional settings, \textsc{FlashSAC} achieves performance comparable to PPO.
        \textbfs{(b) State-based, High DoF:} Dexterous manipulation and humanoid locomotion tasks from IsaacLab and MuJoCo Playground. In high-dimensional settings, \textsc{FlashSAC} substantially outperforms PPO in both asymptotic return and wall-clock efficiency.
        \textbfs{(c) Sim-to-Real:} Humanoid locomotion on the Unitree G1 platform. \textsc{FlashSAC} enables sim-to-real transfer within minutes, whereas PPO requires hours of training.
        }
    \label{fig:teaser}
\end{center}

\section{Introduction}
\label{sec:introduction}

The long-standing goal of robot learning is to develop agents that generalize across a wide range of tasks in the real world.
While large-scale imitation learning from real-world data has recently yielded impressive results in robotic control \cite{kim2025openvla, team2025gemini}, reinforcement learning (RL) from simulation remains a core paradigm when expert demonstrations are unavailable, incomplete, or insufficient \cite{kober2013rl_robotics_survey, arulkumaran2017rl_survey, zhao2020sim2real}.

To date, sim-to-real RL has been most successful in relatively constrained domains such as quadruped locomotion \cite{hwangbo2019leggedsim2real, rudin2022learning} and gripper-based manipulation \cite{andrychowicz2020manisim2real}, which are characterized by low-dimensional state–action spaces and extremely high-throughput simulators \cite{tao2024maniskill3, mittal2025isaaclab, genesis2025Genesis, zakka2025mujoco}. In this regime, on-policy methods such as Proximal Policy Optimization (PPO) \cite{schulman2017ppo, schwarke2025rslrl} have proven effective: PPO is stable, easy to tune, and its data inefficiency is acceptable when fresh on-policy data can be collected cheaply.

However, this regime is becoming less representative of modern robot learning. Emerging applications—including humanoid locomotion \cite{sferrazza2024humanoidbench}, dexterous manipulation \cite{chen2022bidex, wang2024dexteours_pan}, and vision-based control \cite{robocasa2024, jiang2025dexmimicgen}—involve much higher-dimensional state and action spaces, where policy evaluation and improvement from narrowly distributed on-policy data become substantially harder. Simultaneously, simulation grows increasingly expensive due to complex contact dynamics \cite{lin2021softgym, wang2025dexgarmentlab}, and larger policy architectures further raise rollout costs \cite{kim2025openvla, team2025gemini}. In this setting, repeatedly discarding past experience in favor of freshly collected data becomes increasingly inefficient in both sample complexity and wall-clock time.

Off-policy RL offers a natural alternative. By reusing diverse experience from a replay buffer, off-policy methods can achieve substantially higher data efficiency than on-policy approaches \cite{mnih2015human, lillicrap2015ddpg, haarnoja2018sac}. This advantage is particularly appealing in high-dimensional robotic tasks, where broader data coverage can support better policy evaluation and improvement. Yet despite this promise, off-policy RL has not become the default choice for sim-to-real transfer, as it often suffers from slow training and instability \cite{fujimoto2018td3, li2023offpolicyoverfit}.

A central challenge is learning an accurate value function from broad replay data. Off-policy methods train a critic $Q_\theta$ by minimizing a bootstrapped Bellman objective,
\begin{equation}
\label{eq:intro_bellman_loss}
\mathcal{L}_Q
=
\mathbb{E}_{(s,a,r,s') \sim \mathcal{D}}
\left[
\left(
Q_\theta(s,a)
-
\left(r + \gamma Q_\theta(s',a')\right)
\right)^2
\right]
,
\end{equation}
where transitions $(s,a,r,s')$ are sampled from a replay buffer $\mathcal{D}$ and $a' \sim \pi(\cdot \mid s')$. 
In high-dimensional settings, fitting this critic accurately over diverse replay data often requires many gradient updates, which not only increases training time but also compounds estimation errors through repeated bootstrapping, as the critic is optimized toward targets that depend on its own predictions \cite{sutton1998reinforcement, van2018deepdeadlytriad}.

In this paper, we present \textsc{FlashSAC}, a fast and stable off-policy RL algorithm built on Soft Actor-Critic. Motivated by scaling trends in supervised learning, \textsc{FlashSAC} sharply reduces the number of gradient updates while compensating with larger models and higher data throughput, improving training efficiency and better matching modern large-scale simulation pipelines. However, larger critics can further exacerbate instability under bootstrapping. To maintain stability, \textsc{FlashSAC} explicitly controls critic update dynamics by bounding weight, feature, and gradient norms, thereby preventing the accumulation of critic errors.

We evaluate \textsc{FlashSAC} on more than 60 locomotion and manipulation tasks across 10 simulators, spanning high-dimensional state-based control, vision-based control, and sim-to-real humanoid locomotion. Across this benchmark suite, \textsc{FlashSAC} consistently outperforms PPO and strong off-policy baselines in both final performance and training efficiency, with the largest gains on the most challenging tasks such as dexterous manipulation and humanoid locomotion. 
In sim-to-real humanoid walking, \textsc{FlashSAC} reduces training time from hours to minutes while maintaining stable real-world deployment,
demonstrating that off-policy RL can be both fast and stable for scalable sim-to-real robot learning.


\section{Related Work}

\subsection{On-Policy Reinforcement Learning}

On-policy RL has been the dominant paradigm for simulation-based robot learning when environment interaction is cheap and massively parallelizable.
Among on-policy methods, PPO~\cite{schulman2017ppo} is particularly popular for its stability, ease of implementation, and robustness to hyperparameter choices.
Combined with modern high-throughput simulators~\cite{makoviychuk2021isaacgym, tao2024maniskill3, mittal2025isaaclab, genesis2025Genesis}, PPO has enabled successful sim-to-real transfer in relatively constrained domains such as quadruped locomotion~\cite{hwangbo2019leggedsim2real, rudin2022learning, schwarke2025rslrl} and rigid-body manipulation~\cite{andrychowicz2020manisim2real}.
Its widespread adoption has motivated a line of work improving upon it, such as stronger trust-region guarantees~\cite{xie2024simple}, better exploration via expanding action space~\cite{liao2026gpo}, and lower gradient variance via pathwise estimates~\cite{voelcker2025reppo}.

However, on-policy methods fundamentally rely on freshly collected data and discard experience generated by earlier policies~\cite{sutton1998reinforcement}.
As task dimensionality increases, achieving sufficient state–action coverage via on-policy rollouts becomes increasingly expensive.
While importance sampling can, in principle, correct for policy mismatch and enable data reuse~\cite{shelton2001importance, mahmood2014weighted_importance}, importance weights in high-dimensional continuous action spaces exhibit extremely high variance, rendering this strategy impractical in modern robotic learning.

\subsection{Off-Policy Reinforcement Learning}

Off-policy RL decouples data collection from policy optimization by storing transitions in a replay buffer and reusing them across updates~\cite{sutton1998reinforcement, mnih2015human}.
This is especially appealing in high-dimensional robotic tasks, where diverse experience supports better policy evaluation than narrowly distributed on-policy rollouts.

\subsubsection*{Off-Policy Model-Based RL} 

Model-based RL further improves sample efficiency by learning environment dynamics and using them for planning or imagined rollouts~\cite{sutton1990mbrl, janner2019mbpo}.
Recent approaches such as DreamerV3~\cite{hafner2023dreamerV3} and TD-MPC2~\cite{hansen2024tdmpc2} have demonstrated strong performance in vision-based domains by planning in learned latent spaces.
However, learning accurate dynamics models and performing repeated planning procedures significantly increases per-step training cost~\cite{moreland2023mbrl_survey}, often limiting their scalability in wall-clock critical settings.

\subsubsection*{Off-Policy Model-Free RL}

Model-free off-policy algorithms such as DDPG~\cite{lillicrap2015ddpg}, TD3~\cite{fujimoto2018td3}, and SAC~\cite{haarnoja2018sac} learn policies and value functions directly from replayed experience without explicit dynamics models. Their simplicity and data reuse make them attractive for robotic control. However, as discussed in Section~\ref{sec:introduction}, off-policy model-free RL suffers from three persistent challenges: \emph{slow training}, \emph{unstable training dynamics}, and \emph{exploration in high-dimensional action spaces}.
The first two challenges stem from the bootstrapped Bellman objective illustrated in Equation~\ref{eq:intro_bellman_loss}: fitting a critic over diverse replay data in high-dimensional state-action spaces requires many gradient updates, directly increasing training time. Because critic targets depend on the critic's own predictions, approximation and extrapolation errors at poorly supported state-action pairs compound across updates~\cite{sutton1998reinforcement, van2018deepdeadlytriad}. The third arises because the maximum-entropy formulation of SAC alone is often insufficient to maintain coherent exploration in high-dimensional action spaces~\cite{eberhard2023pink_noise, hollenstein2022ou_noise}, motivating dedicated noise mechanisms and exploration schemes.

Prior work has primarily addressed each challenge in isolation. To improve \emph{speed}, one line of work scales data throughput via parallel simulation and large replay buffers~\cite{li2023pql, raffin2025sac_part1, raffin2025sac_part2, seo2025fasttd3, seo2025fastsac, obando2025simplicial}. For example, FastTD3~\cite{seo2025fasttd3} and FastSAC~\cite{seo2025fastsac} achieve strong wall-clock efficiency in humanoid locomotion but relies on small networks (${\sim}$0.2M parameters), which limits its asymptotic performance. Scaling to larger networks is difficult in this setting, as increased model capacity exacerbates instability under bootstrapped training.

To improve \emph{stability}, a second line of work constrains value-function sensitivity by bounding feature, weight, and gradient norms~\cite{bhatt2019crossq, lee2024plastic, gallici2024pqn, nauman2024bro, lee2024simba, lee2025simbaV2, lyle2024nap, palenicek2025CrossQWN, palenicek2025xqc, nauman2025brc, elsayed2024weightclipping, hussing2024dissecting, vasan2024avg}, or by other measures such as reinitialization~\cite{nikishin2022primacy, sokar2023dormant}, distillation~\cite{vincent2025eau}, ensembling~\cite{chen2021redq, hiraoka2021dropoutrl}, and alternative critic target networks~\cite{vincent2025bridging, hendawy2025minto}. These constraints limit error amplification under distribution shift and repeated bootstrapping, enabling training with larger networks that achieve higher asymptotic performance. However, the increased model capacity requires more gradient updates to converge, resulting in slower training in data-rich simulation regimes.

To improve \emph{exploration}, a third line of work exploits off-policy RL's ability to decouple data collection from policy optimization, injecting temporally-correlated action noise to obtain coherent trajectories such as Ornstein--Uhlenbeck processes~\cite{hollenstein2022ou_noise}, pink noise~\cite{eberhard2023pink_noise}, parameter-space noise~\cite{plappert2018parameterspace}, state-dependent exploration~\cite{ruckstiess2008sde, raffin2020gsde}, and temporally-extended action repetition~\cite{dabney2020ez_greedy}. However, exploration mechanisms alone leave the bottlenecks of slow and unstable training untouched, limiting their impact as model capacity and data throughput scale.

\textsc{FlashSAC} unifies these directions: it achieves fast training by sharply reducing gradient updates while scaling model capacity and data throughput, maintains stable training dynamics by jointly bounding weight, feature, and gradient norms, and adopts a lightweight noise-repetition scheme that produces temporally-correlated exploration without per-environment overhead.

\section{Preliminary}

In this section, we introduce the RL framework and algorithmic foundation upon which \textsc{FlashSAC} is built.

\subsection{Markov Decision Process (MDP)}
We model robotic control as a discounted Markov Decision Process (MDP),
\(
\mathcal{M}=(\mathcal{S}, \mathcal{A}, P, r, \gamma)
\),
where \(\mathcal{S}\) denotes the state space, \(\mathcal{A}\) denotes the continuous action space, 
\(P(s'|s,a)\) denotes the transition dynamics, \(r(s,a)\) denotes the reward function, 
and \(\gamma \in [0,1)\) is the discount factor.

At each timestep \(t\), the agent observes \(s_t \in \mathcal{S}\), samples an action \(a_t \in \mathcal{A}\), receives a reward \(r_t=r(s_t,a_t)\), and transitions to the next state \(s_{t+1} \sim P(\cdot \mid s_t,a_t)\). The goal is to learn a policy \(\pi(a|s)\) that maximizes the discounted sum of rewards.

\subsection{Soft Actor Critic (SAC)}
\textsc{FlashSAC} builds upon SAC~\cite{haarnoja2018sac}, a widely used off-policy RL algorithm.
SAC stores transitions \((s, a, r, s')\) collected under past policies in a replay buffer \(\mathcal{D}\), and trains the policy using samples drawn from this buffer.

Beyond maximizing expected return, SAC incorporates an entropy regularization term that encourages exploration.
This entropy maximization is particularly important in high-dimensional state–action spaces, where insufficient exploration can lead to poor coverage of the replay buffer and exacerbate approximation and extrapolation errors.

To reduce approximation errors in bootstrapped value learning, SAC commonly employs clipped double Q-learning \cite{fujimoto2018td3}, maintaining two action-value functions \(Q_{\phi_1}(s,a)\) and \(Q_{\phi_2}(s,a)\).
The minimum of the two estimates is used when forming targets, reducing the impact of optimistic value errors.

Concretely, the policy \(\pi_\theta(a|s)\) is optimized by minimizing
\begin{equation}
\mathcal{L}_{\pi}(\theta)
=
\mathbb{E}_{s \sim \mathcal{D},\, a \sim \pi_\theta}
\big(
\alpha \log \pi_\theta(a|s)
-
\min_{i=1,2} Q_{\phi_i}(s,a)
\big),
\label{eq:policy_objective}
\end{equation}
where \(\alpha > 0\) controls the relative importance of entropy.

Each critic is trained by minimizing a bootstrapped Bellman error using slowly updated target networks \(\bar{\phi}_1\) and \(\bar{\phi}_2\), which are updated via exponential moving average:
\begin{equation}
\bar{\phi}_j \leftarrow \tau \phi_j + (1 - \tau)\bar{\phi}_j, \; j \in \{ 1, 2 \}
\end{equation}
where \(\tau \in (0,1)\) is the target update rate.

For \(i \in \{1,2\}\), the critic weights \(\phi_i\) are optimized by minimizing the Bellman loss
\begin{equation}
\mathcal{L}_{Q}(\phi_i)
=
\mathbb{E}_{(s,a,r,s') \sim \mathcal{D}}
\left[
Q_{\phi_i}(s,a) - y
\right]^2,
\end{equation}
where the target value is
\begin{equation}
y =
r + \gamma \big(
\min_{j=1,2} Q_{\bar{\phi}_j}(s',a')
- \alpha \log \pi_\theta(a'|s')
\big),
a' \sim \pi_\theta(\cdot|s').
\end{equation}

\section{FlashSAC}

\textsc{FlashSAC} is a fast and stable off-policy RL algorithm for high-dimensional robotic control.
It achieves strong asymptotic performance with fast wall-clock time through three complementary mechanisms:
(i) fast training by scaling data and model while reducing gradient updates (\S\ref{method:fast_training}),
(ii) stable training by constraining critic update dynamics (\S\ref{method:stable_training}), and
(iii) broad exploration for diverse data coverage (\S\ref{method:exploration}).

\subsection{Fast Training}
\label{method:fast_training}

On-policy methods such as PPO discard all collected data after each iteration.
In high-dimensional robotic tasks where simulation is expensive, this data inefficiency becomes a critical bottleneck.
Off-policy RL reuses past experience from a replay buffer, but conventionally requires many gradient updates per transition to extract sufficient learning signal, which slows wall-clock time and compounds bootstrapping errors.

\textsc{FlashSAC} takes a different approach inspired by the scaling trends observed in supervised learning: under a fixed compute budget, larger models trained with larger batches and fewer updates converge faster than smaller models with frequent updates~\cite{kaplan2020scaling}.
This principle has been difficult to apply in off-policy RL, because increased model capacity tends to amplify critic instability under bootstrapping.
\textsc{FlashSAC} resolves this tension by stabilizing critic training through constrained update dynamics (\S\ref{method:stable_training}), enabling a regime of high data throughput, large models, and infrequent gradient updates.

\subsubsection*{Massively Parallel Simulation}
 
We collect data using $1024$ parallel simulation environments, enabling rapid accumulation of diverse trajectories.
While many off-policy RL setups rely on a small number of environments~\cite{haarnoja2018sac, fujimoto2018td3}, high-throughput data collection is critical for maintaining adequate coverage of the state-action space in high-dimensional tasks.

\subsubsection*{Large-Capacity Replay Buffer}

\textsc{FlashSAC} uses a replay buffer of up to 10M transitions, an order of magnitude larger than the 1M commonly used in standard off-policy configurations~\cite{lee2024simba, palenicek2025xqc}.
In high-dimensional tasks, rare but important state-action pairs can be easily overwritten in smaller buffers, leading to catastrophic forgetting and inducing extrapolation error.
A larger buffer preserves such long-tail experiences and maintains the diversity of training data available to the critic throughout learning~\cite{fedus2020replay}.
 
\subsubsection*{Large Model, Large Batch, Fewer Updates}
 
Standard off-policy RL baselines use small MLPs (0.2-0.5M parameters, 2-3 layers) to avoid instability~\cite{haarnoja2018sac, seo2025fasttd3}. 
In contrast, \textsc{FlashSAC} employs a 2.5M-parameter, 6-layer network for both the actor and critic, paired with a batch size of 2048 that nearly saturates GPU utilization.
The updates-to-data ratio is set to 2/1024, meaning only 2 gradient updates are performed per 1024 new transitions.
Although such infrequent updates are typically ineffective in off-policy RL, the combination of large batches, higher learning rates, and increased model capacity enables fast convergence with fewer updates.

\subsubsection*{Code Optimization}

\textsc{FlashSAC} is implemented in PyTorch~\cite{paszke2019pytorch}, with both training and inference JIT-compiled to minimize Python overhead.
We use mixed-precision throughout training~\cite{micikevicius2017mixed_precision}, which reduces wall-clock time by 5-10\%.

\subsection{Stable Training}
\label{method:stable_training}

Scaling data and model accelerates training but does not prevent instability arising from bootstrapped critic updates.
In the Bellman backup, estimation errors at next-state action pairs propagate into the current Q-value targets and can be recursively amplified through repeated updates.
This problem worsens with both state-action dimensionality and model capacity, making stability a prerequisite for scaling.
\textsc{FlashSAC} addresses this by constraining weight, feature, and gradient norms throughout training via the following mechanisms.

\begin{figure}[t]
\begin{center}
\includegraphics[width=0.52\textwidth]{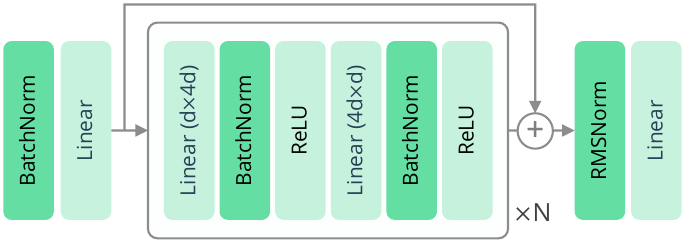}
\end{center}
\caption{\textbfs{\textsc{FlashSAC} Architecture.} The architecture consists of stacked inverted residual blocks with pre-activation batch normalization and post-RMS normalization.}
\label{figure:architecture}
\end{figure}

\subsubsection*{Inverted Residual Backbone}
 
The backbone stacks inverted residual blocks inspired by the Transformer feedforward block~\cite{vaswani2017attention} (Figure~\ref{figure:architecture}).
Each block expands features to a higher dimension via an inverted bottleneck~\cite{howard2017mobilenets}, projects back to the original dimension, and adds a residual connection~\cite{he2016resnet} to stabilize gradient propagation.
After the final block, we apply RMSNorm~\cite{zhang2019rmsnorm} to bound per-sample feature norms before value heads, preventing out-of-distribution inputs from producing unbounded activations that destabilize bootstrapping.

\subsubsection*{Pre-activation Batch Normalization}
 
Replay data is collected by a mixture of evolving policies, inducing non-stationary input distributions.
Without normalization, feature activations can saturate (e.g., dead ReLUs~\cite{abbas2023crelu}), degrading gradient flow~\cite{dohare2023maintaining, lyle2023understanding_plasticity, lee2024slow}.
We apply batch normalization~\cite{ioffe2015batchnorm} before each nonlinearity to keep activations well-scaled.
We choose batch normalization over layer normalization~\cite{ba2016layernorm} because it exploits large-batch statistics from diverse replay data, yielding a smoother loss landscape with a lower effective condition number~\cite{santurkar2018batchnorm_why, bhatt2019crossq, palenicek2025xqc}.

\subsubsection*{Cross-Batch Value Prediction}
 
Batch normalization computes statistics per batch, so the predicted Q-values and target Q-values receive different normalization when computed in separate forward passes.
Following~\cite{bhatt2019crossq}, we concatenate current and next-state transitions into a single batch so that both share the same statistics, ensuring consistency in the Bellman update.
 
\subsubsection*{Distributional Critic with Adaptive Reward Scaling}
Following~\cite{lee2025simbaV2, palenicek2025xqc}, we represent the Q-value as a categorical distribution over $n_\text{atom}$ atoms uniformly spaced on $[G_{\min}, G_{\max}]$. The network predicts atom probabilities and is trained via cross-entropy loss against the projected Bellman target~\cite{bellemare2017c51}.
This distributional formulation smooths the optimization landscape and reduces sensitivity to noisy targets~\cite{palenicek2025xqc}.
 
To keep returns within the distributional critic's fixed support, we normalize rewards directly rather than centering returns~\cite{naik2024reward_centering} or scaling losses~\cite{schaul2021td_scaling}.
We track the running discounted return variance $\sigma^2_{t,G}$ and maximum magnitude $G_{t,\max}$, and scale as:
\begin{equation}
    \bar{r}_t = \frac{r_t}{\max\!\left(\sqrt{\sigma_{t,G}^2 + \epsilon},\; G_{t, \max} / G_{\max} \right)}.
\end{equation}
This bounds effective returns while maintaining a consistent scale throughout training.

\subsubsection*{Weight Normalization}
 
Uncontrolled weight growth increases Q-value variance and amplifies estimation errors under bootstrapping~\cite{lyle2024nap}.
After each gradient step, we project each weight vector onto the unit-norm sphere~\cite{van2017bw_norm, loshchilov2024ngpt, lee2025simbaV2, palenicek2025CrossQWN, palenicek2025xqc} and each normalization parameter vector $(\gamma, \beta)$ to norm $\sqrt{d}$.
This constrains the network to encode information through direction rather than scale.

\subsection{Exploration}
\label{method:exploration}

Off-policy RL can decouple data collection from policy optimization, allowing exploration strategies that pursue broad state-action coverage independently of the current policy.
\textsc{FlashSAC} employs two complementary mechanisms.

\subsubsection*{Unified Entropy Target}

Maximum-entropy RL with automatic temperature tuning~\cite{haarnoja2018sac} encourages sustained exploration, but requires specifying a target entropy.
Standard practice sets this target per task, which is impractical across embodiments with varying action dimensions.
We instead parameterize the target entropy via a fixed action standard deviation $\sigma_{\text{tgt}}$.
For a Gaussian policy with diagonal covariance, this gives:
\begin{equation}
    \bar{\mathcal{H}} = \tfrac{1}{2} |\mathcal{A}| \log \left( 2 \pi e \, \sigma_{\text{tgt}}^2 \right),
\end{equation}
which scales linearly with action dimension, ensuring consistent exploration across embodiments without per-task tuning.
We set $\sigma_{\text{tgt}} = 0.15$ in all experiments.

\subsubsection*{Noise Repetition}
 
Temporally correlated action noise is commonly used to improve exploration in sparse-reward settings, with pink noise~\cite{eberhard2023pink_noise} and Ornstein--Uhlenbeck noise~\cite{hollenstein2022ou_noise} being widely used. However, these methods are ill-suited to massively parallel simulations, as they require per-environment correlated-noise processes, which incur substantial computational and memory overhead.
 
We propose \emph{Noise Repetition}, a lightweight alternative that induces temporal correlation using minimal local state.
At each repetition interval, a noise vector $\epsilon \sim \mathcal{N}(0, I)$ is sampled for action selection and held constant for $k$ consecutive steps.
The repetition length $k$ is drawn from a Zeta distribution with probability mass function $P(k) \propto k^{-s}$~\cite{dabney2020ez_greedy}, favoring short repeat intervals while occasionally producing long, correlated action sequences.

\section{Experiments}

We evaluate \textsc{FlashSAC} on a diverse suite of robotic control tasks, measuring both asymptotic performance and wall-clock time (measured on a single RTX 5090 GPU). 
Our experiments span low- and high-dimensional state-based control, vision-based control, and sim-to-real humanoid locomotion.

\begin{figure*}[t!]
\centering
\begin{center}
\includegraphics[width=0.999\textwidth]{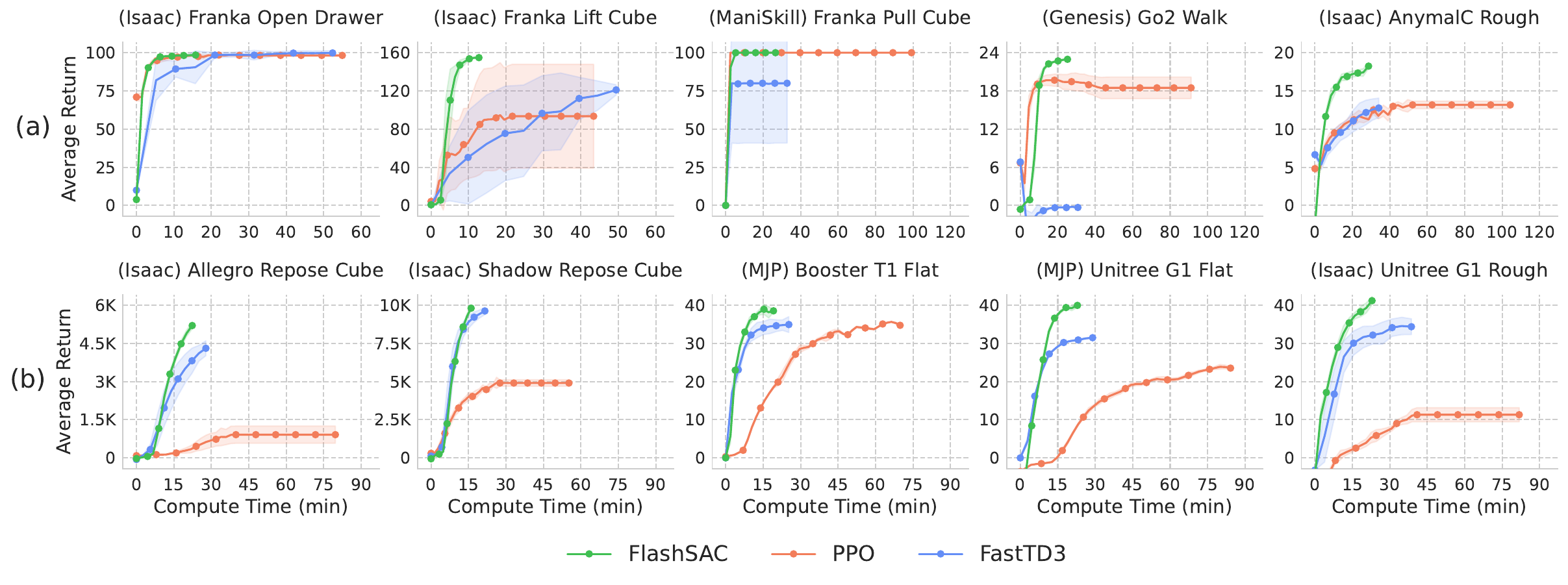}
\end{center}
\vspace{-2mm}
\caption{\textbfs{Results on State-Based RL, GPU-based Simulators.} Learning curves on select tasks from IsaacLab~\cite{mittal2025isaaclab}, ManiSkill~\cite{tao2024maniskill3}, Genesis~\cite{genesis2025Genesis}, and MuJoCo Playground~\cite{zakka2025mujoco}. We evaluate performance on \textbfs{(a)} low-dimensional tasks with gripper manipulation and quadruped locomotion, and \textbfs{(b)} high-dimensional tasks involving dexterous manipulation and humanoid locomotion. While \textsc{FlashSAC} is comparable to PPO~\cite{schwarke2025rslrl} in low-dimensional tasks, \textsc{FlashSAC} significantly outperforms PPO in high-dimensional tasks.}
\label{figure:state_based_result_gpu}
\end{figure*}

\subsection{State-Based RL on GPU-based Simulators}
\label{subsection:state_rl_gpu}

\subsubsection*{Experimental Setup} 

We evaluate on 25 state-based control tasks drawn from four GPU-based simulators: IsaacLab~\cite{mittal2025isaaclab}, MuJoCo Playground~\cite{zakka2025mujoco}, ManiSkill3~\cite{tao2024maniskill3}, and Genesis~\cite{genesis2025Genesis}, all of which enable large-scale sample collection at minimal wall-clock cost.

The tasks span a wide range of state–action dimensionalities:
\begin{itemize}[leftmargin=1.5em]
\item Low-dim (15 tasks): Gripper-based manipulation (Franka) and quadruped locomotion (AnyMal-C/D, Unitree Go2).
\item High-dim (10 tasks): Dexterous manipulation (Allegro, Shadow Hand) and humanoid locomotion (Unitree G1, H1, Booster T1).
\end{itemize}
A complete task list is provided in \S~\ref{appendix:environment_details}.

We compare \textsc{FlashSAC} against strong, widely adopted baselines:
\begin{itemize}[leftmargin=1.5em]
    \item PPO~\cite{schwarke2025rslrl}: A highly optimized on-policy implementation from RSL-RL, representative of current best practices in sim-to-real robotic RL.
    \item FastTD3~\cite{seo2025fasttd3}: A wall-clock–optimized off-policy method designed for high throughput simulations.
\end{itemize}
Whenever available, we report published results; otherwise, we reproduce results using official implementations.

Off-policy methods (\textsc{FlashSAC} and FastTD3) are trained for 50M environment steps.
To probe asymptotic performance, PPO is trained for 200M steps, requiring approximately $3\times$ the compute of \textsc{FlashSAC}.
While baseline methods use task-specific hyperparameter tuning, \textsc{FlashSAC} is evaluated using a single unified configuration across all tasks, varying only the discount factor $\gamma$ to match simulator defaults (e.g., $0.99$ for IsaacLab, $0.97$ for Playground).

\subsubsection*{Experimental Results} Figure~\ref{figure:state_based_result_gpu} summarizes performance on representative tasks, with full results in \S~\ref{appendix:full_results}.

On low-dimensional tasks, \textsc{FlashSAC} slightly outperforms PPO (Figure~\ref{figure:state_based_result_gpu}.a). As consistent with prior findings, on-policy methods remain effective when state–action spaces are small, and simulation throughput is high enough to collect a large volume of samples.

On high-dimensional tasks, \textsc{FlashSAC} demonstrates a clear and consistent advantage (Figure~\ref{figure:state_based_result_gpu}.b). Across dexterous manipulation and humanoid locomotion benchmarks, \textsc{FlashSAC} converges reliably to higher asymptotic performance while requiring substantially less wall-clock time than PPO.

Compared to FastTD3, \textsc{FlashSAC} is markedly more stable, converging across all tasks where FastTD3 frequently fails or underperforms (e.g., Go2Walk, Franka Pull Cube).
When both methods converge, \textsc{FlashSAC} achieves higher asymptotic performance, with the largest gains observed in humanoid locomotion, where larger model capacity is particularly beneficial.

\begin{figure*}[t!]
\centering
\begin{center}
\includegraphics[width=0.999\textwidth]{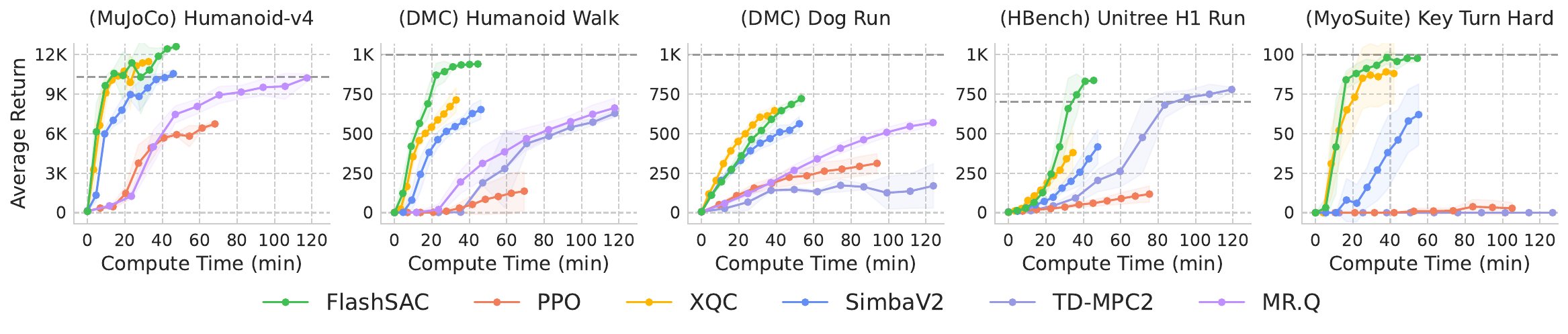}
\end{center}
\caption{\textbfs{Results on State-Based RL, CPU-based Simulators.} Learning curves on select tasks from MuJoCo~\cite{towers2024gymnasium, todorov2012mujoco}, DMC~\cite{tassa2018dmc}, HumanoidBench~\cite{sferrazza2024humanoidbench} and MyoSuite~\cite{caggiano2022myosuite}.
We primarily evaluate high-dimensional tasks, involving dexterous manipulation and humanoid locomotion. \textsc{FlashSAC} significantly outperforms PPO, as well as strong off-policy RL and model-based RL baselines in both compute efficiency and asymptotic performance.
}
\label{figure:state_based_result_cpu}
\end{figure*}

\subsection{State-Based RL on CPU-based Simulators}
\label{subsection:state_rl_cpu}

\subsubsection*{Experimental Setup} 

We further evaluate \textsc{FlashSAC} on 40 single-environment, CPU-based continuous-control tasks drawn from four established benchmarks: MuJoCo~\cite{todorov2012mujoco}, DeepMind Control Suite~\cite{tassa2018dmc}, MyoSuite~\cite{caggiano2022myosuite}, and HumanoidBench~\cite{sferrazza2024humanoidbench}.
Unlike GPU-based simulators, these benchmarks use a single environment instance, placing greater emphasis on sample efficiency rather than wall-clock throughput.

We compare \textsc{FlashSAC} against strong sample-efficient baselines:
\begin{itemize}[leftmargin=1.5em]
    \item PPO~\cite{schwarke2025rslrl}: A highly optimized on-policy implementation from RSL-RL, included to assess whether on-policy methods remain viable in the low-sample regime.
    \item XQC~\cite{palenicek2025xqc}: A recent off-policy method coupled with batch-normalization designed for high sample efficiency.
    \item SimbaV2~\cite{lee2025simbaV2}: An improved variant of Simba~\cite{lee2024simba} for stable off-policy learning.
    \item TD-MPC2~\cite{hansen2024tdmpc2}: A model-based method that combines off-policy RL with model-predictive planning.
    \item MR.Q~\cite{fujimoto2025towards}: A recent model-free method using a model-based objective for better representation learning.
\end{itemize}

As in the GPU-based setting, \textsc{FlashSAC} uses a single unified configuration across all tasks, with only minimal adjustments to match each benchmark's conventions.
Since sample collection is considerably slower with a single environment, the CPU-based configuration differs from the GPU setting by reducing the batch size from 2048 to 512 and setting the update-to-data ratio to 1, reflecting the lower data throughput.

\subsubsection*{Experimental Results}

As illustrated in Figure~\ref{subsection:state_rl_cpu}, \textsc{FlashSAC} consistently outperforms all baselines across representative tasks in this sample-efficient regime. Full per-task results appear in \S~\ref{appendix:full_results}. PPO performs particularly poorly here, as on-policy methods cannot reuse experience and thus suffer under limited sample budgets. 

These results confirm that the design choices in \textsc{FlashSAC} generalize beyond massively parallel GPU-based simulation: even in the classical single-environment setting, where sample efficiency is the primary bottleneck, \textsc{FlashSAC} matches or exceeds dedicated sample-efficient methods without task-specific tuning.

\subsection{Vision-based RL}
\label{sec:vision_based_rl}

\subsubsection*{Experimental Setup} 

We extend our evaluation to vision-based control, where high rendering cost and low environment throughput severely limit the number of transitions collected per unit time, making data efficiency critical. We evaluate on 8 tasks from the DMControl Suite~\cite{tunyasuvunakool2020dm_control}, spanning manipulation and mono/bi-pedal locomotion. A complete task list is provided in \S~\ref{appendix:environment_details}.

Given the low throughput of visual environments, we focus on off-policy baselines:
\begin{itemize}[leftmargin=1.5em]
\item DrQ-v2~\cite{yarats2021drqv2}: A DDPG-based~\cite{lillicrap2015ddpg} method that improves data efficiency through image augmentation~\cite{laskin2020rad}.
\item MR.Q~\cite{fujimoto2025towards}: An off-policy method that incorporates a dynamics modeling objective to improve representation learning.
\end{itemize}

As in the CPU-based state experiments (\S~\ref{subsection:state_rl_cpu}), sample collection is slow; we reuse the same hyperparameters from that setup, adapting only the following to match the standard DrQ-v2 configuration~\cite{yarats2021drqv2}: (i) a lightweight convolutional encoder (3 convolutional layers followed by a linear bottleneck), (ii) frame stacking of the three most recent frames (84 × 84 × 9) for temporal reasoning without recurrent architectures, and (iii) 3-step returns for better credit assignment. All methods are trained for 1M environment steps with an action repeat of 2. Full details are in \S~\ref{appendix:environment_details}.

\begin{figure*}[t!]
\centering
\begin{center}
\includegraphics[width=0.99\textwidth]{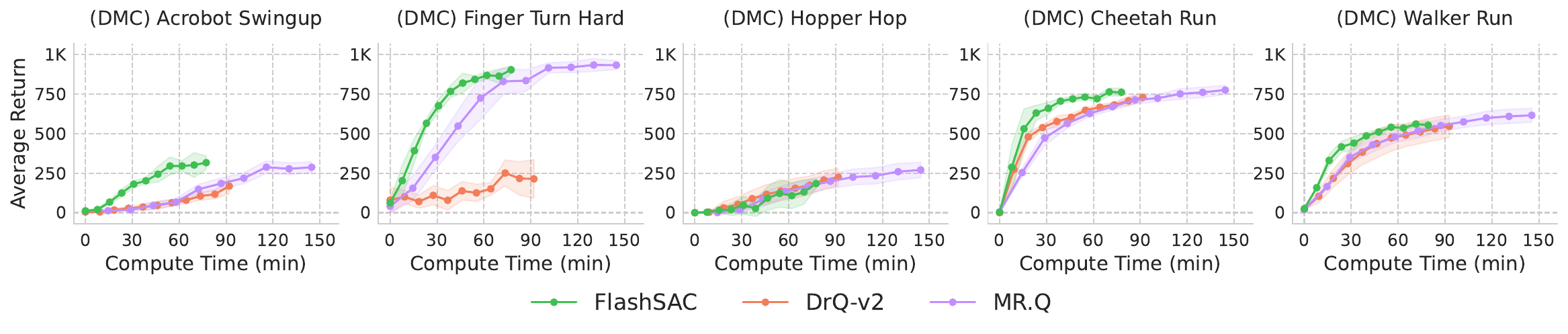}
\end{center}
\vspace{-3mm}
\caption{\textbfs{Results on Vision-Based RL.} Learning curves on selected tasks from vision-based DMControl Suite~\cite{tassa2018dmc}. We assess learning performance in low-dimensional environments, including pendulum manipulation and bipedal locomotion. \textsc{FlashSAC} achieves better compute efficiency and higher asymptotic performance.}
\label{figure:vision_based_result}
\end{figure*}

\noindent
\subsubsection*{Experimental Results}

Figure~\ref{figure:vision_based_result} shows representative learning curves, with full results in \S~\ref{appendix:full_results}. Across tasks, \textsc{FlashSAC} matches or exceeds all baselines in asymptotic performance while converging faster in wall-clock time. DrQ-v2 is sample-efficient but unstable, failing to converge in several environments (e.g., Finger Turn Hard). MR.Q achieves high final performance but incurs additional computational cost from its auxiliary dynamics model. In contrast, \textsc{FlashSAC} achieves competitive or superior results with a single set of hyperparameters, without task-specific exploration or auxiliary objectives. 

We note that the stabilization techniques in \textsc{FlashSAC} are orthogonal to such extensions; for example, MR.Q's representation learning objective could be layered on top for further gains in visual feature learning.

\subsection{Sim-to-Real Transfer}

Off-policy RL is often regarded as unreliable for sim-to-real transfer, particularly in high-dimensional systems, where training instability can lead to unsafe behaviors~\cite{mock2023comparison}. We evaluate whether \textsc{FlashSAC} enables reliable sim-to-real transfer on a challenging 29-DoF Unitree G1 humanoid performing blind locomotion. 

\subsubsection*{Experimental Setup} 

We train blind locomotion policies in simulation using a terrain curriculum comprising pyramid stairs, discrete grids, waves, and pits. The curriculum consists of 10 terrain levels with stair heights ranging from 
0 to 23cm (step width 32cm, platform width 3m). Terrain difficulty is increased automatically using a game-inspired curriculum~\cite{rudin2022walkminutes}. To facilitate sim-to-real transfer, we apply large-scale domain randomization alongside the terrain curriculum; full details are provided in \S~\ref{appendix:details_sim2real}.

As a baseline, we use PPO with the sim-to-real pipeline of~\cite{nahrendra2023dreamwaq}. \textsc{FlashSAC} adopts the same sim-to-real adaptation techniques for a fair comparison. Both methods use implicit system identification via a context estimator~\cite{nahrendra2023dreamwaq} and an asymmetric actor–critic formulation~\cite{pinto2017asymmetric}, where the critic receives privileged information (e.g., contact states and height maps) during training. Both methods share identical reward design and coefficients, combining velocity tracking with regularization terms penalizing foot slip, excessive torque, action discontinuities, and orientation instability. \textsc{FlashSAC} uses the same architecture and hyperparameters as in the state-based experiments~\ref{subsection:state_rl_cpu}.

\subsubsection*{Experimental Results} 

On flat terrain (Figure~\ref{fig:teaser}.(c)), \textsc{FlashSAC} achieves stable real-world locomotion after approximately 20 minutes of training, whereas PPO requires about 3 hours to reach comparable performance. The learned policy supports omnidirectional locomotion (forward, backward, and lateral) without re-training.

\begin{figure*}[t!]
\centering
\begin{center}
\includegraphics[width=0.999\textwidth]{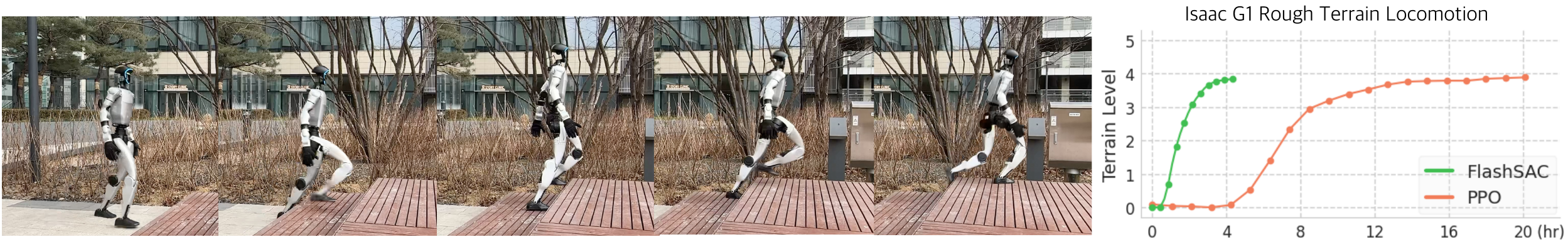}
\end{center}
\vspace{-2mm}
\caption{\textbfs{Sim-to-real Stair Climbing on Unitree G1.} \textsc{FlashSAC} achieves stable real-world stair climbing after only 4 hours of training on simulation, whereas PPO requires nearly 20 hours to reach the same capability.}
\label{figure:result_sim2real}
\end{figure*}

The advantage of \textsc{FlashSAC} is more pronounced on rough terrain (Figure~\ref{figure:result_sim2real}). In the real-world setup, the robot faces stairs of 15cm height, 60cm width, and 1.5m platform width—conditions unseen during training, which uses different stair dimensions. \textsc{FlashSAC} successfully climbs these stairs after approximately 4 hours of training, while PPO requires nearly 20 hours to achieve a similar capability. 

Overall, \textsc{FlashSAC} reduces the training time for sim-to-real humanoid locomotion by nearly an order of magnitude compared to PPO while maintaining stable and safe behaviors.

\section{Analysis}

In this section, we analyze the factors underlying \textsc{FlashSAC}'s performance across four aspects. We first examine how off-policy learning yields broader state–action coverage than on-policy methods~(\S\ref{subsection:on_policy_vs_off_policy}). We then investigate three design choices central to \textsc{FlashSAC}: scaling data collection and model capacity for faster training~(\S\ref{subsection:exp_faster_training}), architectural ablations that improve training stability~(\S\ref{subsection:exp_stable_training}), and the effect of entropy and temporal correlation on exploration~(\S\ref{subsection:exploration}). All experiments are conducted in four IsaacLab environments~\cite{mittal2025isaaclab}: cube reorientation with the Allegro and Shadow Hands, and flat and rough terrain locomotion with the G1.

\begin{figure}[h]
\centering
\begin{center}
\includegraphics[width=0.55\textwidth]{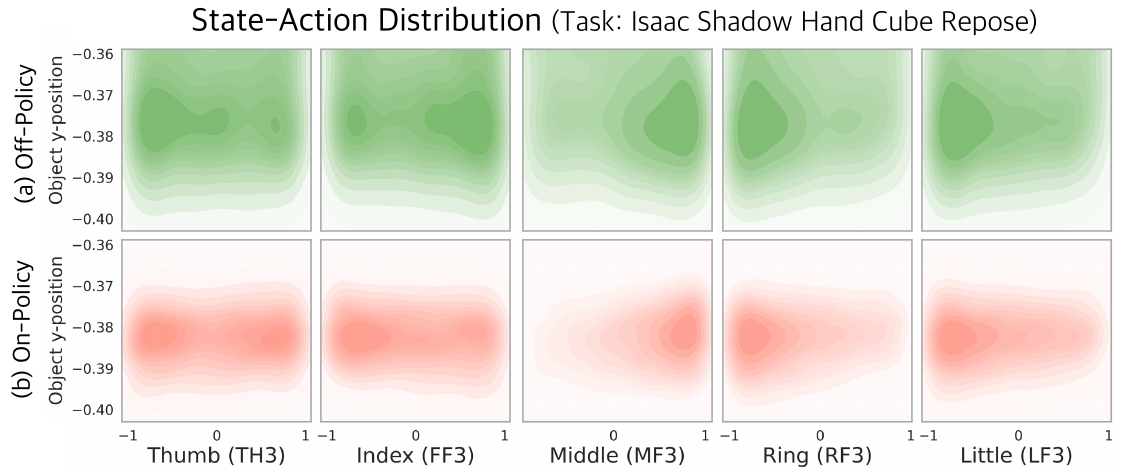}
\end{center}
\vspace{-1mm}
\caption{\textbfs{Off-Policy vs. On-Policy Data Coverage.} 
We train \textsc{FlashSAC} for 1M steps on the Shadow Hand cube reorientation task. 2D density plots show the joint distribution of object y-position and fingertip joint actions for \textbfs{(a) off-policy} samples from the replay buffer and \textbfs{(b) on-policy} samples collected by rolling out the final policy for an additional 1M steps. Off-policy data covers a substantially broader region of the state–action space.}
\label{figure:off_vs_on}
\end{figure}

\subsection{Off-Policy vs On-Policy}
\label{subsection:on_policy_vs_off_policy}

We train \textsc{FlashSAC} for 1M steps on the IsaacLab Shadow Hand task with a replay buffer of size 1M, then collect 1M additional on-policy transitions by rolling out the final policy. 

Figure~\ref{figure:off_vs_on} compares the state–action coverage of the two datasets via 2D density plots of finger actions versus object y-position. 
Off-policy data (Figure~\ref{figure:off_vs_on}.(a)) covers a substantially broader region of the state–action space, reflecting experience accumulated across diverse behavior policies stored in the replay buffer. On-policy data (Figure~\ref{figure:off_vs_on}.(b)), by contrast, is tightly concentrated around the final policy's distribution. This disparity suggests that limited state–action coverage is a key factor in the reduced effectiveness of on-policy methods on high-dimensional tasks, where achieving comparable coverage would require substantially more data collection.

\subsection{Scaling Ablation for Faster Training}
\label{subsection:exp_faster_training}

We study how scaling data, model capacity, and reducing the number of gradient updates~(\S\ref{method:fast_training}) affects the compute efficiency of \textsc{FlashSAC}. We perform univariate ablations over five hyperparameters: batch size, replay buffer size, network width, network depth, and update-to-data (UTD) ratio. 

\begin{figure*}[t!]
\centering
\begin{center}
\includegraphics[width=0.999\textwidth]{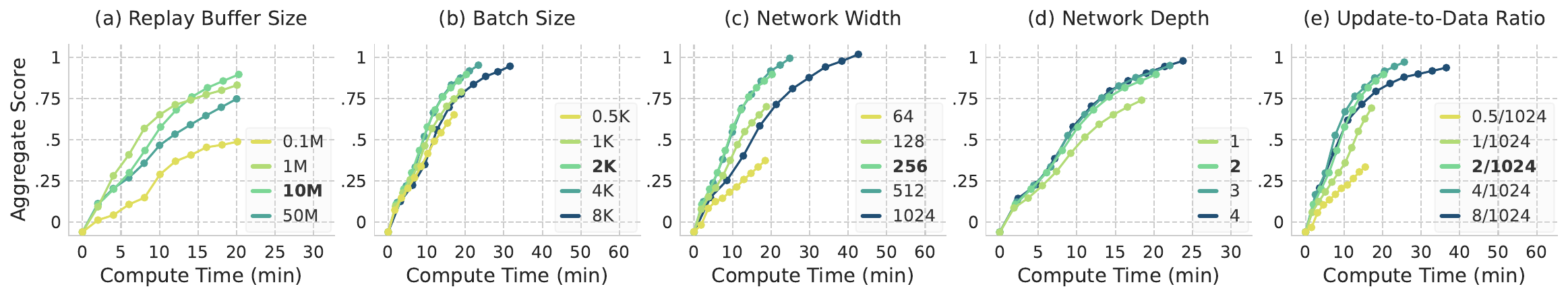}
\end{center}
\vspace{-2mm}
\caption{\textbfs{Scaling Results.} \textbfs{(a)} Replay buffer size trades off training stability and efficiency. \textbfs{(b–e)} Scaling batch size, and model capacity while reducing the UTD ratio accelerates convergence.}
\label{figure:ablation_fast}
\end{figure*}

Figure~\ref{figure:ablation_fast} shows learning curves plotted against wall-clock time. Increasing replay buffer size improves performance up to 10M transitions by stabilizing training (Figure~\ref{figure:ablation_fast}.(a)). However, overly large buffers (e.g., 50M) slow learning because recent high-quality samples are drawn less frequently, though they can achieve slightly higher asymptotic performance given sufficient training time. 

Figures~\ref{figure:ablation_fast}.(b)–(e) exhibit trends consistent with established scaling laws~\cite{kaplan2020scaling}: increasing batch size and model capacity, along with reducing the UTD ratio, accelerate convergence. Most existing off-policy RL methods rely on small architectures for training stability (e.g., width 128 with inverted bottlenecks and block depth 1), which limits convergence speed. The scaling mechanisms of \textsc{FlashSAC} enable higher-capacity models, resulting in substantially faster convergence.

\begin{figure*}[t!]
\centering
\begin{center}
\includegraphics[width=0.999\textwidth]{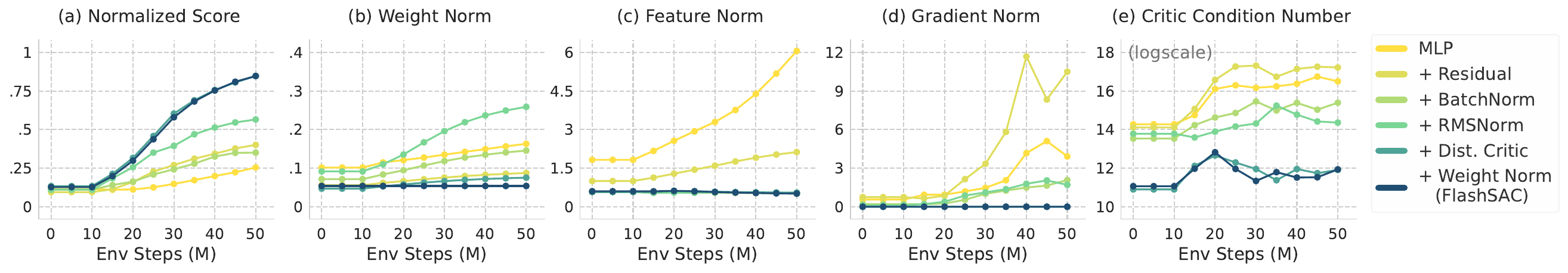}
\end{center}
\vspace{-2mm}
\caption{\textbfs{Architectural Ablations.}  Starting from a standard MLP, each component is incrementally added to build up to \textsc{FlashSAC}. Each addition stabilizes training by constraining weight, feature, and gradient norms while reducing the condition number.}
\label{figure:ablation_stable}
\end{figure*}

\subsection{Architectural Ablation for Stable Training}
\label{subsection:exp_stable_training}
We analyze the contribution of each architectural component in \textsc{FlashSAC} (\S\ref{method:stable_training}) to determine whether the proposed design stabilizes training.
Beyond final task performance, we measure parameter, feature, and gradient norms throughout training as indicators of optimization stability. Following \cite{palenicek2025xqc}, we also measure the condition number of the critic loss landscape, where larger values correspond to poorly conditioned updates that can exacerbate critic error amplification.

Starting from a standard MLP critic, we incrementally add: Residual Blocks, Batch Normalization, Post RMSNorm, Distributional Critics with Reward Scaling, and Weight Normalization. Figure~\ref{figure:ablation_stable} summarizes the results. As components are added, parameter, feature, and gradient norms remain bounded throughout training with no uncontrolled growth, indicating well-behaved critic updates and reduced error amplification. The condition number also decreases monotonically, reaching its lowest value with the full \textsc{FlashSAC} architecture.

These gains in optimization stability directly translate to improved task performance (Figure~\ref{figure:ablation_stable}.(a)), underscoring the importance of controlling update dynamics in off-policy RL. While weight normalization alone yields modest gains, it improves robustness in sample-limited regimes and is therefore retained in the final design.

\subsection{Exploration Ablation}
\label{subsection:exploration}

We analyze \textsc{FlashSAC}'s exploration strategy~(\S\ref{method:exploration}): unifying the entropy target $\sigma_{tgt}$ and noise repetition.

\begin{figure}[t]
\centering
\begin{center}
\includegraphics[width=0.5\textwidth]{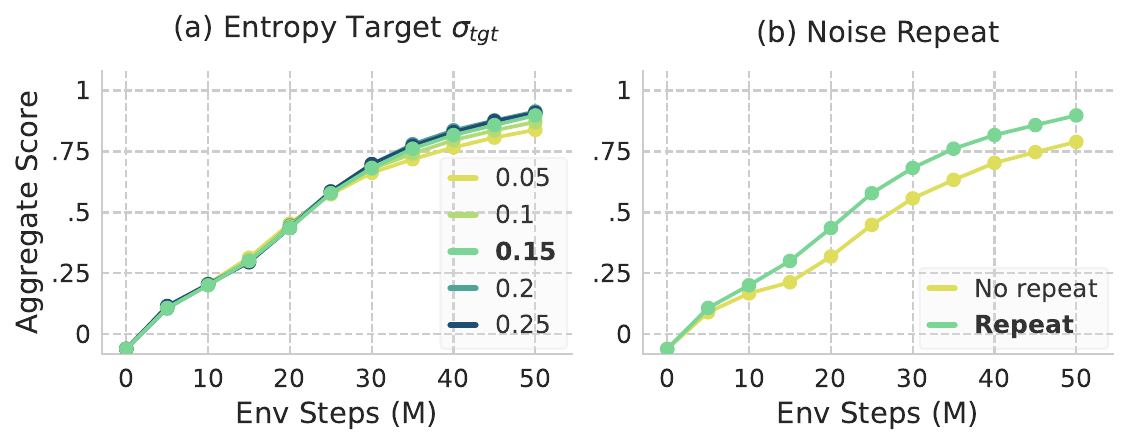}
\end{center}
\vspace{-1mm}
\caption{\textbfs{Exploration Ablation Results.} \textbfs{(a) Unified entropy target:} The optimal entropy target $\sigma_{tgt}$lies in the range 0.15 to 0.2 across tasks, enabling a unified setting without task-specific tuning. \textbfs{(b) Noise repetition:} Repeating action noise accelerates convergence and improves asymptotic performance.}
\label{figure:ablation_exploration}
\end{figure}

Figure~\ref{figure:ablation_exploration}.(a) shows the effect of varying the entropy target $\sigma_{tgt}$  across values \{0.05,0.1,0.15,0.2,0.25\}. Performance is largely insensitive to this hyperparameter, with all settings converging to similar asymptotic scores. This robustness simplifies tuning in practice, as the default value ($\sigma_{tgt}$=0.15) performs well without task-specific adjustment.

Figure~\ref{figure:ablation_exploration}.(b) compares training with and without noise repetition. Disabling noise repeat leads to slower convergence and lower aggregate scores, confirming that temporally correlated exploration is crucial for \textsc{FlashSAC}. Repeating sampled action noise across consecutive steps produces coherent exploratory trajectories rather than uncorrelated perturbations that are quickly averaged out by the dynamics in high-dimensional control tasks.

\section{Lessons and Opportunities}

We presented \textsc{FlashSAC}, a fast and stable off-policy RL framework for high-dimensional robotics. As the robotics community moves toward high-dimensional~\cite{gu2025humanoid_survey}, perception-rich~\cite{robocasa2024, jiang2025dexmimicgen}, and contact-intensive tasks~\cite{wang2025dexgarmentlab}, the scalability of on-policy RL becomes increasingly constrained. Off-policy RL is an appealing alternative, but its adoption has been limited by slow training speed and instability in critic learning arising from function approximation error and bootstrapped updates. \textsc{FlashSAC} addresses these challenges through two complementary mechanisms: scaling data and model capacity while reducing the number of gradient updates for faster training, and integrating explicit architectural constraints on critic updates for stable optimization. Together, these yield strong asymptotic performance and up to an order-of-magnitude reduction in wall-clock time compared to on-policy methods.

Stabilized off-policy learning opens new opportunities for robot learning. Improved data efficiency makes it feasible to train larger policies, incorporate vision and other rich sensory inputs~\cite{su2024sim2real_tactile}, and leverage slower but more realistic simulators~\cite{puig2023habitat}. Off-policy methods also naturally support learning from a mixture of demonstrations and self-collected experience~\cite{ball2023rlpd}. While this work focuses on state-based control, extending these critic-stabilization principles to tactile-based learning is a promising direction for future work.

\section*{Acknowledgements}

We would like to express our gratitude to Younggyo Seo and Yekyung Nah for their valuable feedback on this paper.
This work was supported by the Institute for Information \&
communications Technology Planning \& Evaluation (IITP)
grant funded by the Korea government (MSIT) (RS-2019-
II190075, Artificial Intelligence Graduate School Program
(KAIST)).
This research was also funded by the research cluster ``Third Wave of AI'', funded by the excellence program of the Hessian Ministry of Higher Education, Science, Research and the Arts, hessian.AI and by the Deutsche Forschungsgemeinschaft (DFG, German Research Foundation) under Germany's Excellence Strategy (EXC-3057/1 ``Reasonable Artificial Intelligence'', Project No. 533677015).
It was further partially supported by the German Federal Ministry of Research, Technology and Space (BMFTR) under the Robotics Institute Germany (RIG).

\clearpage
\bibliographystyle{assets/plainnat}
\bibliography{paper}

\clearpage
\newpage
\beginappendix

\section{Measuring Wall-Clock Time}
\label{appendix:estimating_time}

We adopt a unified protocol for measuring wall-clock time across all environments and simulators. All experiments are executed on identical hardware (AMD Ryzen 9 9950X3D CPU with an RTX 5090 GPU).

We decompose the total runtime into two components: (i) \emph{environment interaction time}, which consists of environment stepping, physics simulation, and replay buffer operations; and (ii) \emph{algorithm update time}, which consists of policy inference and gradient-based parameter updates.

Environment interaction time is measured independently for each environment. For a fixed environment, this cost is determined by the underlying simulator and is independent of the learning algorithm.

Algorithm update time is measured separately using representative benchmark environments. Specifically, we profile update time on MuJoCo Humanoid-v4 for state-based experiments and on DMC Walker-run for vision-based experiments. This update cost is reused across all environments, as it is largely determined by the algorithm’s architecture and optimization procedure rather than the environment itself.

The total wall-clock time for a given environment is estimated by summing the measured environment interaction time and the corresponding algorithm update time. While differences in observation and action dimensionality across environments may lead to small variations in update cost, these effects are negligible in practice and do not affect the overall comparison.

\section{Environment Details}
\label{appendix:environment_details}

We evaluate \textsc{FlashSAC} across a diverse set of simulation environments, grouped by simulator. For each environment, we report the observation space, action space, and the score used for normalization.

\subsection{IsaacLab}

We use IsaacLab v2.1.0~\cite{mittal2025isaaclab}, a GPU-accelerated simulation platform built on NVIDIA Isaac Sim. Our evaluation includes 12 tasks spanning gripper manipulation, dexterous hand manipulation, quadruped locomotion, and humanoid locomotion. Scores are normalized per environment using a near-asymptotic performance reference obtained by running \textsc{FlashSAC} for an extended training duration. Full environment specifications are provided in Table~\ref{tab:isaaclab_environment}.

\subsection{Mujoco Playground}

We evaluate four humanoid locomotion tasks from the MuJoCo Playground suite (v0.0.5)~\cite{zakka2025mujoco}, which emphasizes robust whole-body control under complex contact dynamics. For consistency across tasks, normalized scores are uniformly scaled to 40, corresponding to the asymptotic performance observed in converged training runs. Task details are listed in Table~\ref{tab:mjp_environment}.

\subsection{ManiSkill}

We employ ManiSkill~\cite{tao2024maniskill3}, a large-scale benchmark built on the SAPIEN physics engine that supports diverse object geometries and physical interactions. We evaluate six rigid-body manipulation tasks using a gripper-based robotic arm. To ensure strict reproducibility, we use the environment snapshot corresponding to commit hash \texttt{aad75f2}. The full task list is shown in Table~\ref{tab:maniskill_environment}.

\subsection{Genesis}

We benchmark three environments from the Genesis simulator (v0.3.13)~\cite{genesis2025Genesis}, a general-purpose physics engine. We adapt the source code to comply with the standard Gymnasium API. Additional environment details are provided in Table~\ref{tab:genesis_environment}.

\subsection{Gym - MuJoCo}
\label{appendix:environments_gym}

We use the Gym~\citep{towers2024gymnasium} continuous control benchmark simulated with MuJoCo~\citep{todorov2012mujoco}. Our evaluation focuses on five locomotion tasks (version \texttt{v4}) involving multi-body dynamics and contact-rich interactions. To enable comparisons across tasks with different reward scales, we normalize scores using TD7 baselines~\citep{fujimoto2023td7}, with reference points defined by the random policy score and the score achieved after 5M training steps (approximating asymptotic performance). The task list is given in Table~\ref{tab:mujoco_environment}.

\subsection{DeepMind Control Suite}
\label{appendix:environments_dmc}

The DeepMind Control Suite (DMC)~\cite{tassa2018dmc} provides a broad collection of continuous control tasks. We evaluate 10 tasks, ranging from low-dimensional sparse-reward environments (e.g., Cartpole, Pendulum) used to assess exploration, to high-dimensional tasks commonly referred to as \texttt{DMC-Hard}, such as Humanoid and Dog. Observation and action space details are reported in Table~\ref{tab:dmc_environment}.

\subsection{HumanoidBench}
\label{appendix:environments_hb}

HumanoidBench~\citep{sferrazza2024humanoidbench} is a high-dimensional benchmark for whole-body control based on the Unitree H1 humanoid robot. We evaluate 14 locomotion tasks that require stable gait generation and balance. All scores are normalized using the success thresholds defined by the benchmark authors. Task specifications and dimensionalities are summarized in Table~\ref{tab:hbench_environment}.

\subsection{MyoSuite}
\label{appendix:environments_myo}

MyoSuite~\citep{caggiano2022myosuite} models human motor control using physiologically accurate musculoskeletal simulations. We evaluate 10 dexterous manipulation tasks involving the elbow, wrist, and hand. Following the benchmark’s taxonomy, tasks are labeled \texttt{easy} when the target goal is fixed and \texttt{hard} when the goal is randomized. The complete set of evaluated tasks is listed in Table~\ref{tab:myosuite_environment}.

{\centering
\par \vspace{10mm}
\input{tables/info_isaaclab}
\par \vspace{10mm}
\input{tables/info_mjp}
\par \vspace{10mm}
\input{tables/info_maniskill}

\par \vspace{10mm}
\input{tables/info_genesis}

\par \vspace{10mm} 
\input{tables/info_mujoco}
\par \vspace{10mm}
\input{tables/info_dmc}
\par \vspace{10mm}
\input{tables/info_hbench}
\par \vspace{10mm}
\input{tables/info_myosuite}
\par \vspace{10mm}
}

\section{Hyperparameters}
\label{appendix:full_hyperparameters}
\input{tables/hyperparam_gpu}
\par \vspace{10mm}
\input{tables/hyperparam_cpu}

\par \vspace{10mm}
\input{tables/hyperparam_visual}
\par \vspace{10mm}


\section{Sim-to-Real Details}
\label{appendix:details_sim2real}

To evaluate the sim-to-real transfer capability of our algorithm, we train blind locomotion policies in simulation and directly deploy them on hardware without additional fine-tuning. This section describes the detailed experimental setup, including the simulation environment, control architecture, terrain curriculum, observation design, and reward formulation.

\subsubsection{Simulation}

We employed NVIDIA IsaacLab~\cite{mittal2025isaaclab} as the simulation platform to train the \textsc{FlashSAC} controller, with training environments built upon the Legged Gym framework~\cite{rudin2022walkminutes}. The agents were trained in $4096$ parallel environments with domain randomization, and training was completed in approximately 4 hours using a single NVIDIA A100 GPU. The resulting policy networks were directly deployed on the physical robot without additional fine-tuning.

\subsubsection{Low-Level Control} 

The policy network outputs target joint positions at a frequency of $50$ Hz, which are subsequently passed to a low-level PD controller running at $200$ Hz. Within the PD controller, these target joint positions are translated into torque commands using proportional ($K_p$) and derivative ($K_d$) gains. To achieve natural and stable motion, we adopt a heuristic PD gain design following~\cite{liao2025beyondmimic}, with the specific parameters listed in Table~\ref{tab:sim2real_pdgain}.

The PD controller and sensor data measurement routines were implemented using a custom Pybind-based interface to connect our Python RL policy loop with the C++ implementation of Unitree SDK to ensure a real-time low-level routine. This interface enables the transmission of target joint positions to the Unitree SDK through governed by a ROS2 node. 

\begin{figure}[h]
\begin{center}
\includegraphics[width=0.5\textwidth]{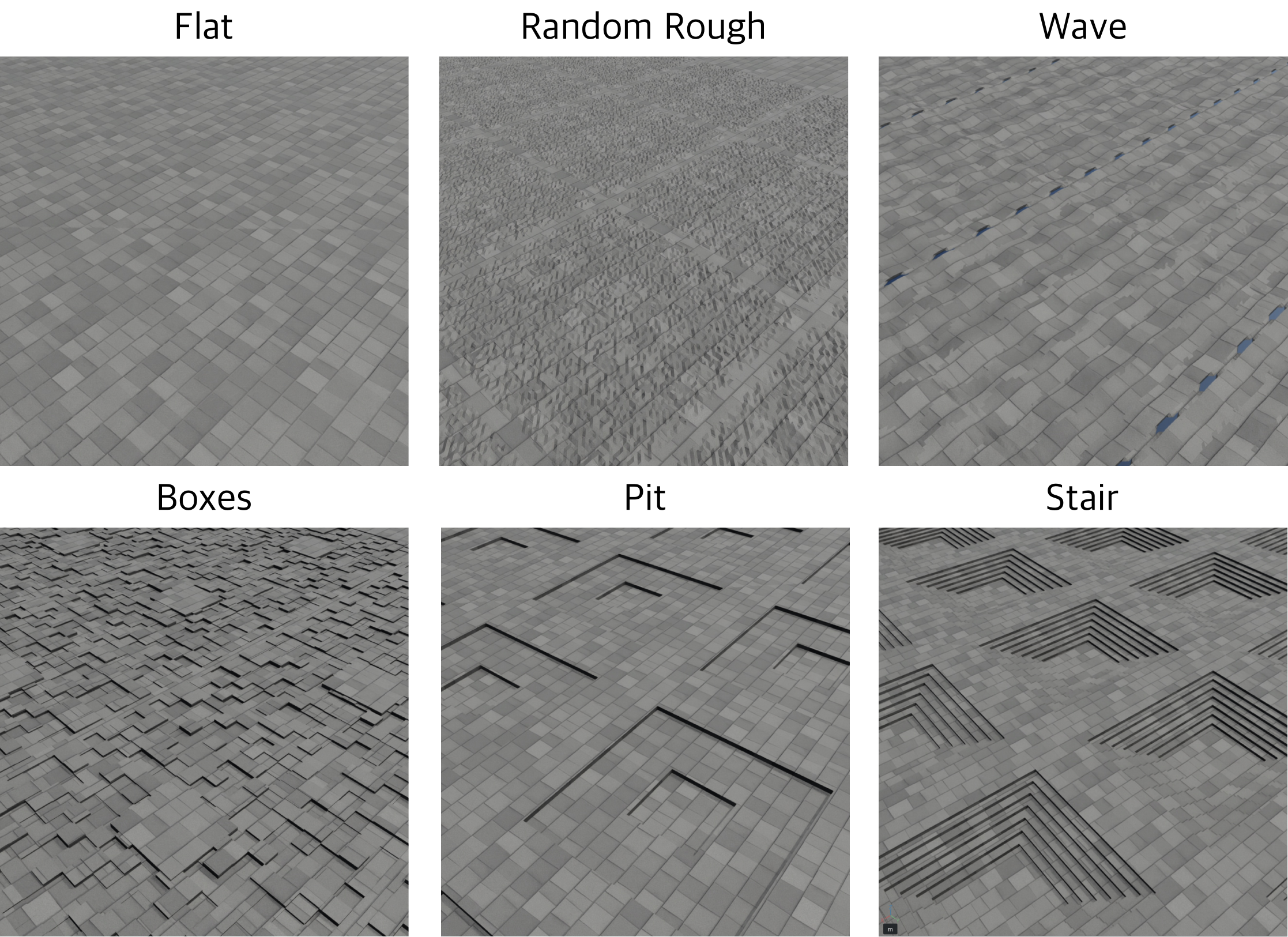}
\end{center}
\caption{\textbf{Terrain configuration.} Blind locomotion policies are trained on flat and diverse rough terrains, including pyramid stairs, random grids, random uniform terrain, wave terrain, and pit terrain. A terrain curriculum progressively increases difficulty across $10$ levels as the policy successfully traverses the environment, enabling stable and adaptive locomotion without exteroceptive perception.}
\label{figure:sim2real_terrain}
\end{figure}

\subsubsection{Terrain Configuration and Curriculum}

We train blind locomotion policies on both flat and rough terrains composed of five terrain types as visualized in Figure~\ref{figure:sim2real_terrain}:

\begin{itemize}[leftmargin=20pt, topsep=1pt, itemsep=0pt]
    \item Pyramid stairs (maximum step height of $23$cm)
    \item Random grids (maximum height of $15$cm)
    \item Random uniform terrain (height noise range of $-2$ to $4$cm)
    \item Wave terrain (maximum amplitude of $20$cm)
    \item Pit terrain (maximum depth of $30$cm)
\end{itemize}

Traversing these terrains without exteroceptive perception requires a high degree of stability and strong in-context adaptation. To facilitate the progressive acquisition of such challenging locomotion skills and to accelerate training, we employ terrain curriculum learning, a widely used technique in legged locomotion~\cite{rudin2022walkminutes, hoeller2024anymal}. The maximum curriculum level is set to 10, and terrain difficulty is automatically increased once the policy successfully traverses 50\% of the environment. We observe that progressing beyond level $5$ is difficult without a perception module and leads to aggressive motions of policies.

\subsubsection{Observation space}

The observation 
\begin{equation}
  \textbf{o}_t=\begin{bmatrix}\boldsymbol{\omega}_t & \textbf{g}_t & \textbf{c}_t & \boldsymbol{q}_t & \boldsymbol{\dot{q}}_t & \textbf{a}_{t-1} \end{bmatrix}^\top,
  \label{eqn:observation_vector}
\end{equation}
includes base angular velocity $\boldsymbol{\omega}_t$, joint position $\boldsymbol{q}_t$, joint velocity $\dot{\boldsymbol{q}}_t$, projected gravity $\boldsymbol{g}_t$, previous action $\boldsymbol{a}_{t-1}$, and velocity command $\boldsymbol{c}_t$, which are accessible on hardware during deployment. In addition, we incorporate the context estimator network (CENet)~\cite{nahrendra2023dreamwaq} into the actors of PPO and \textsc{FlashSAC} to take $\boldsymbol{x}_t = [\boldsymbol{v}_t \; \boldsymbol{o}_t \; \boldsymbol{z}_t]$ as an input. This estimator encodes the proprioceptive observation history $\boldsymbol{o}_{t-H:t-1}$ and outputs predicted base linear velocity $\boldsymbol{v}_t$ and its latents $\boldsymbol{z}_t$, which are further enhanced by auxiliary loss of estimating base linear velocity during training. This history-based encoding enables implicit system identification for sim-to-real transfer and, together with auxiliary velocity estimation, ensures reliable state estimation for stable blind locomotion in real-world settings~\cite{lee2020quadruped, kumar2021rma, ji2022stateest}. Furthermore, the training batch of these observations is augmented via symmetry augmentation~\cite{mittal2024symaug}, where we found that it enhances sample efficiency and produces natural behavior. 
And we adopt an asymmetric actor-critic framework~\cite{pinto2017asymmetric}, where the critic networks of both PPO and \textsc{FlashSAC} also take privileged information, comprising ground-truth base linear velocity, foot contact state, and height map.

\subsubsection{Reward}

In Table~\ref{tab:sim2real_reward}, we summarize the reward configurations used for both algorithms. We adopt a shared reward structure comprising task rewards for base velocity tracking and regularization terms penalizing foot slip, joint torques, action rate, and orientation instability. However, due to the differing learning dynamics between our method and PPO, different reward weights are required for stable real-world deployment~\cite{seo2025fasttd3}. A notable distinction lies in the termination penalty. PPO requires substantial training time to achieve stable locomotion behavior without termination-based shaping, whereas \textsc{FlashSAC}\ does not. Consequently, we apply only a minimal alive reward for \textsc{FlashSAC}\ to avoid premature termination.

\newpage
\input{tables/info_sim2real_pdgain}
\par \vspace{10mm}
\input{tables/info_sim2real_obs}

\par \vspace{10mm}
\input{tables/info_sim2real_reward}
\par \vspace{10mm}
\input{tables/info_sim2real_reward_symbol}
\par \vspace{10mm}

\newpage
\section{Complete Results}
\label{appendix:full_results}

We report the learning curves for each task across all algorithms. The results are plotted against wall-clock time and environment steps to illustrate compute and sample efficiency, respectively.

\subsection{IsaacLab (State-based RL, GPU Simulator)}
\label{appendix:full_results_isaaclab}

\begin{figure}[h!]
    \centering

    \vspace{-0.5em}
    
    \begin{minipage}{\textwidth}
        \centering
        \includegraphics[width=0.89\textwidth]{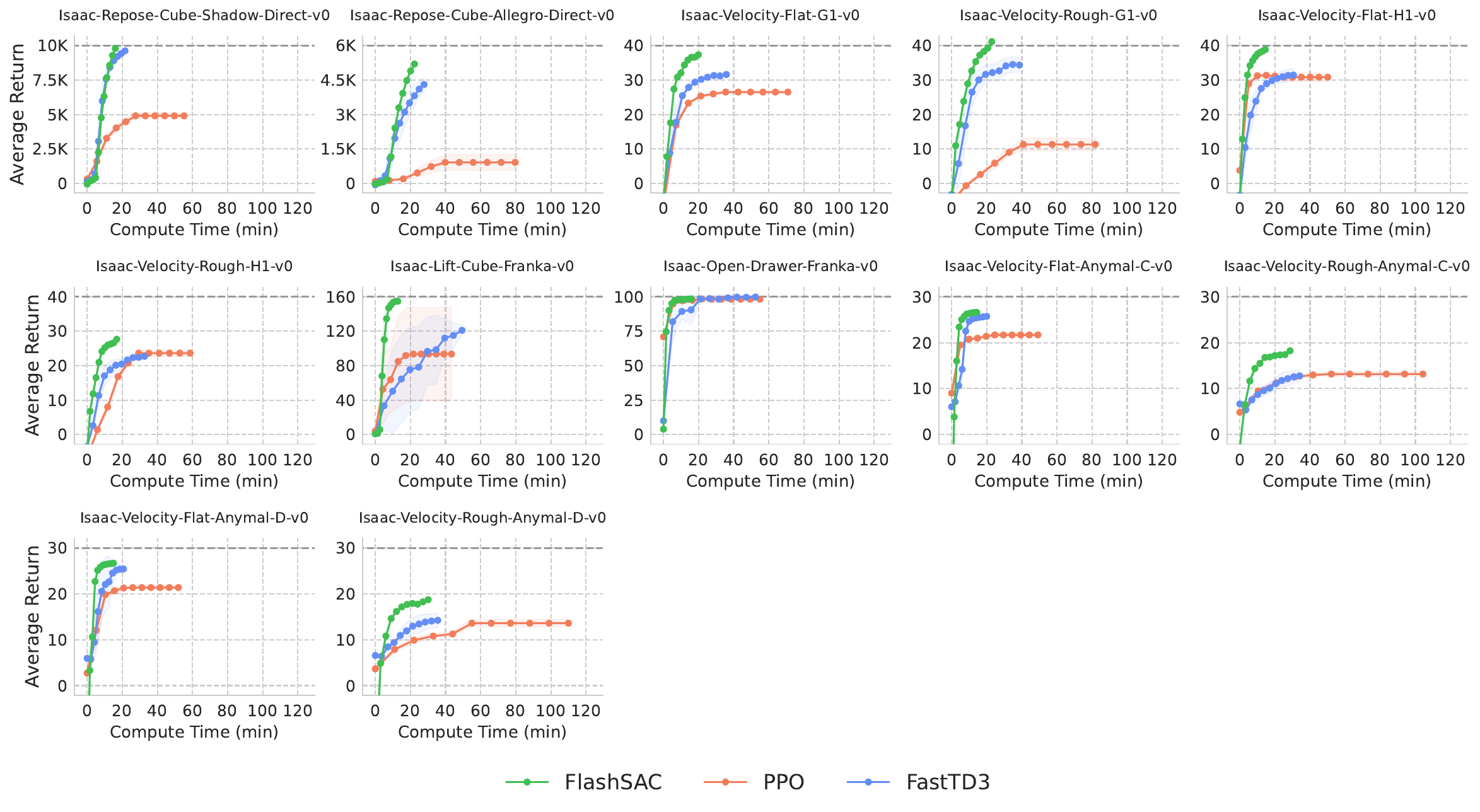}
        \caption{\textbfs{IsaacLab Learning Curves (Compute Efficiency).} Average episode returns in IsaacLab, plotted against total compute time. Results are averaged over random seeds of each algorithm, with shaded regions indicating $95\%$ bootstrap confidence intervals and dotted lines denoting normalize score. All methods are trained for 50M environment steps except for PPO (200M).}
    \end{minipage}
    
    \vspace{1em}
    
    \begin{minipage}{\textwidth}
        \centering
        \includegraphics[width=0.89\textwidth]{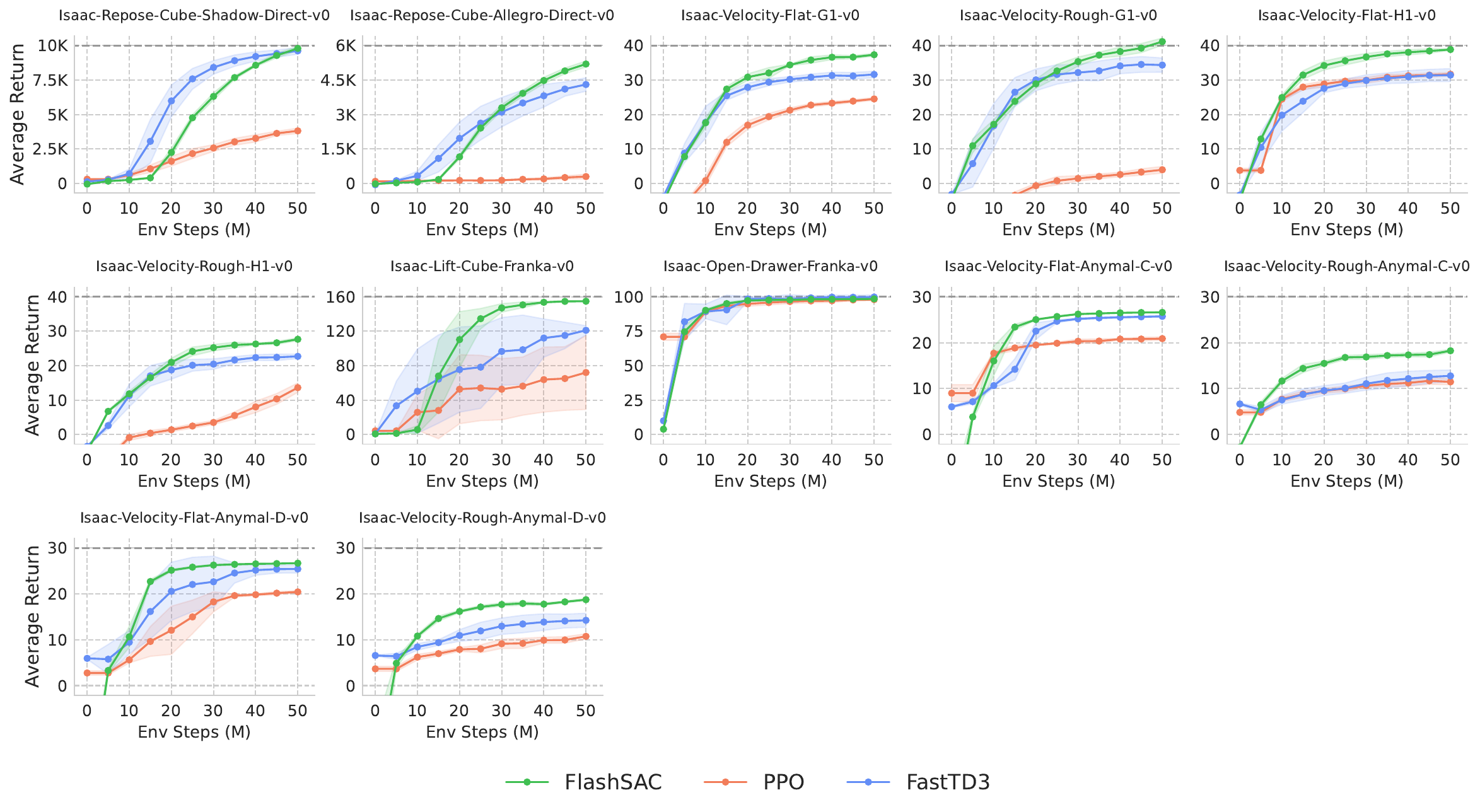}
        \caption{\textbfs{IsaacLab Learning Curves (Sample Efficiency).} Average episode returns in IsaacLab environments, plotted against environment steps. Results are averaged over random seeds of each algorithm, with shaded regions indicating $95\%$ bootstrap confidence intervals and dotted lines denoting normalize score.}
        \label{figure:full_result_isaaclab}
    \end{minipage}
\end{figure}

\clearpage

\subsection{Mujoco Playground (State-based RL, GPU Simulator)}
\label{appendix:full_results_mjp}

\begin{figure}[h!]
    \centering
    \vspace{0.5em}
    
    \begin{minipage}{\textwidth}
        \centering
        \includegraphics[width=0.8\textwidth]{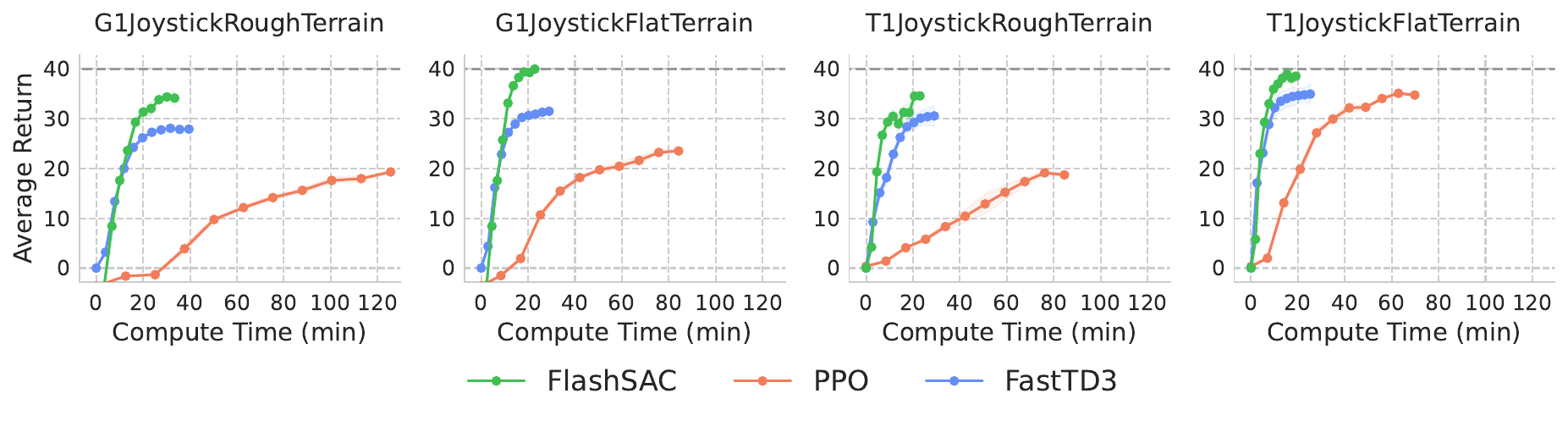}
        \caption{\textbfs{Mujoco Playground Learning Curves (Compute Efficiency).} Average episode returns in Mujoco Playground environments, plotted against total compute time. Results are averaged over random seeds of each algorithm, with shaded regions indicating $95\%$ bootstrap confidence intervals and dotted lines denoting normalize score. All methods are trained for 50M environment steps except for PPO (200M).}
    \end{minipage}
    
    \vspace{3em}
    
    \begin{minipage}{\textwidth}
        \centering
        \includegraphics[width=0.8\textwidth]{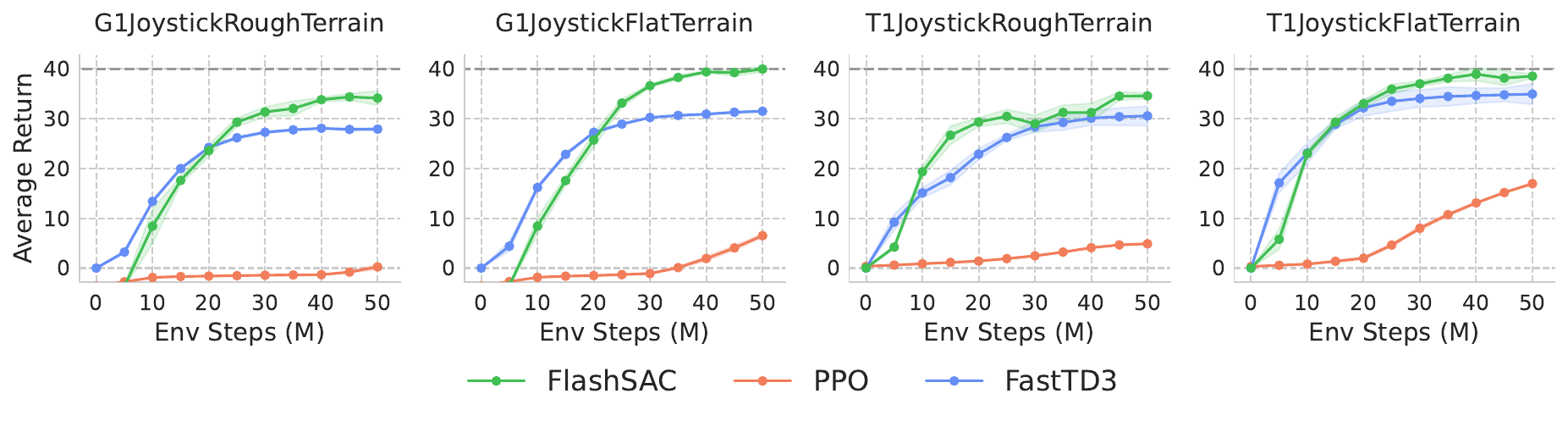}
        \caption{\textbfs{Mujoco Playground Learning Curves (Sample Efficiency).} Average episode returns in Mujoco Playground environments, plotted against environment steps. Results are averaged over random seeds of each algorithm, with shaded regions indicating $95\%$ bootstrap confidence intervals and dotted lines denoting normalize score.}
        \label{figure:full_result_mjp}
    \end{minipage}
\end{figure}

\clearpage

\subsection{ManiSkill (State-based RL, GPU Simulator)}
\label{appendix:full_results_maniskill}

\begin{figure}[h!]
    \centering
    \vspace{0.5em}
    
    \begin{minipage}{\textwidth}
        \centering
        \includegraphics[width=0.99\textwidth]{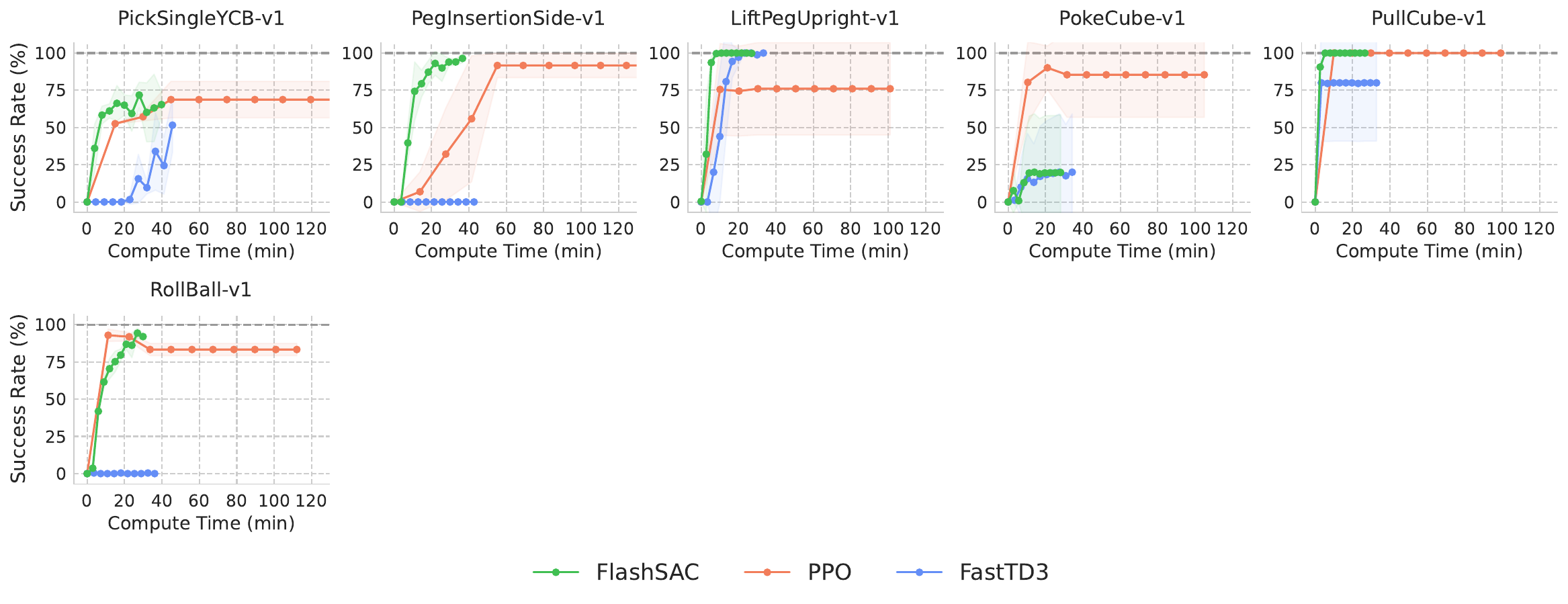}
        \caption{\textbfs{ManiSkill Learning Curves (Compute Efficiency).} Average episode returns in ManiSkill environments, plotted against total compute time.  Results are averaged over random seeds of each algorithm, with shaded regions indicating $95\%$ bootstrap confidence intervals and dotted lines denoting normalized score. All methods are trained for 50M environment steps except for PPO (200M).}
    \end{minipage}
    
    \vspace{3em}
    
    \begin{minipage}{\textwidth}
        \centering
        \includegraphics[width=0.99\textwidth]{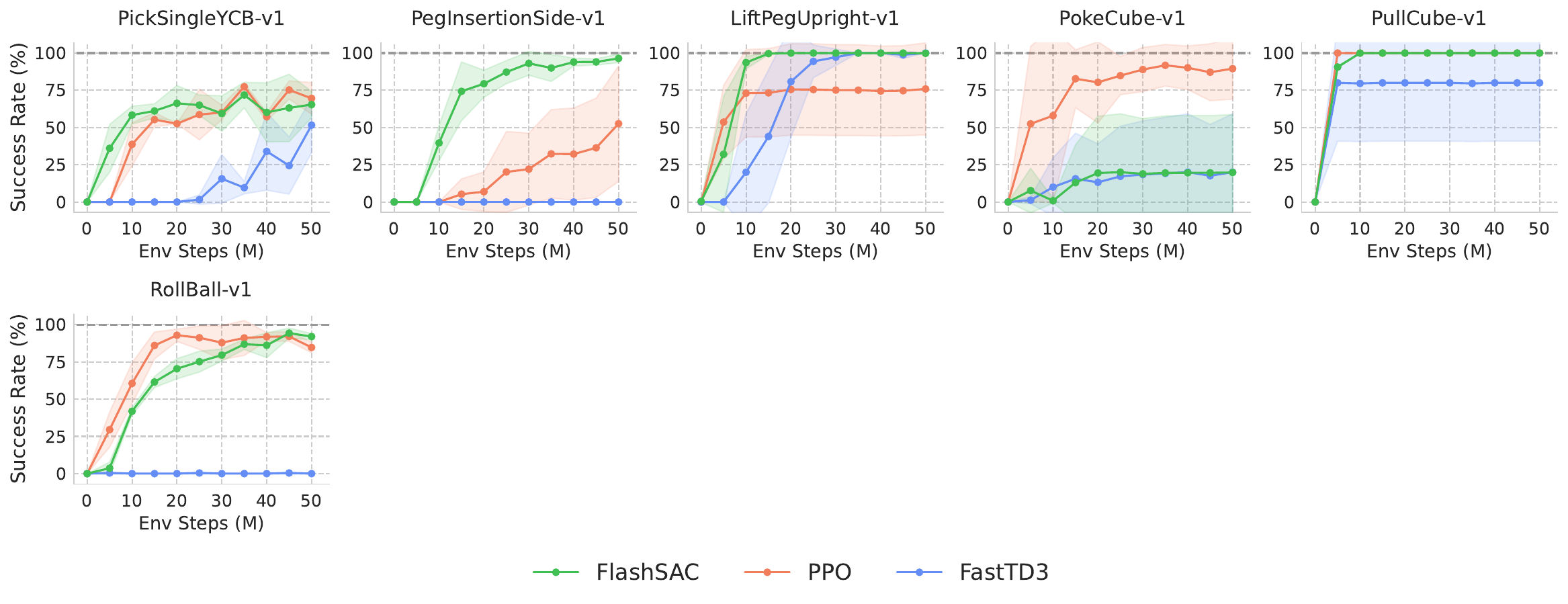}
        \caption{\textbfs{ManiSkill Learning Curves (Sample Efficiency).} Average episode returns in ManiSkill environments, plotted against environment steps. Results are averaged over random seeds of each algorithm, with shaded regions indicating $95\%$ bootstrap confidence intervals and dotted lines denoting normalize score.}
        \label{figure:full_result_maniskill}
    \end{minipage}
\end{figure}

\clearpage

\subsection{Genesis (State-based RL, GPU Simulator)}
\label{appendix:full_results_genesis}

\begin{figure}[h!]
    \centering
    \vspace{0.5em}
    
    \begin{minipage}{\textwidth}
        \centering
        \includegraphics[width=0.65\textwidth]{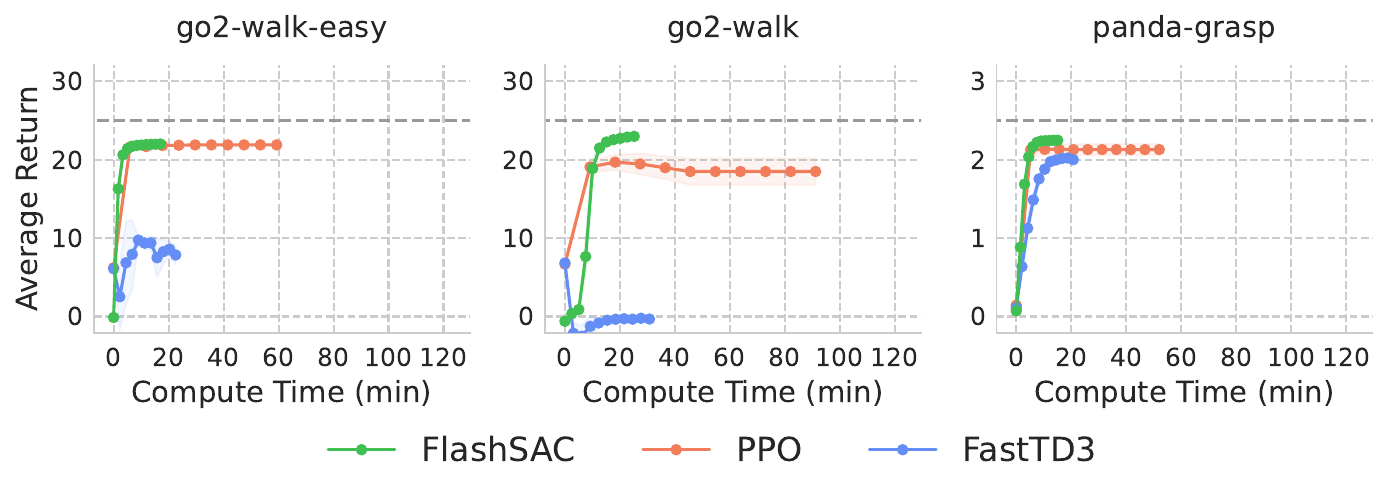}
        \caption{\textbfs{Genesis Learning Curves (Compute Efficiency).} Average episode returns in Genesis environments, plotted against total compute time.  Results are averaged over random seeds of each algorithm, with shaded regions indicating $95\%$ bootstrap confidence intervals and dotted lines denoting normalize score. All methods are trained for 50M environment steps except for PPO (200M).}
    \end{minipage}
    
    \vspace{3em}
    
    \begin{minipage}{\textwidth}
        \centering
        \includegraphics[width=0.65\textwidth]{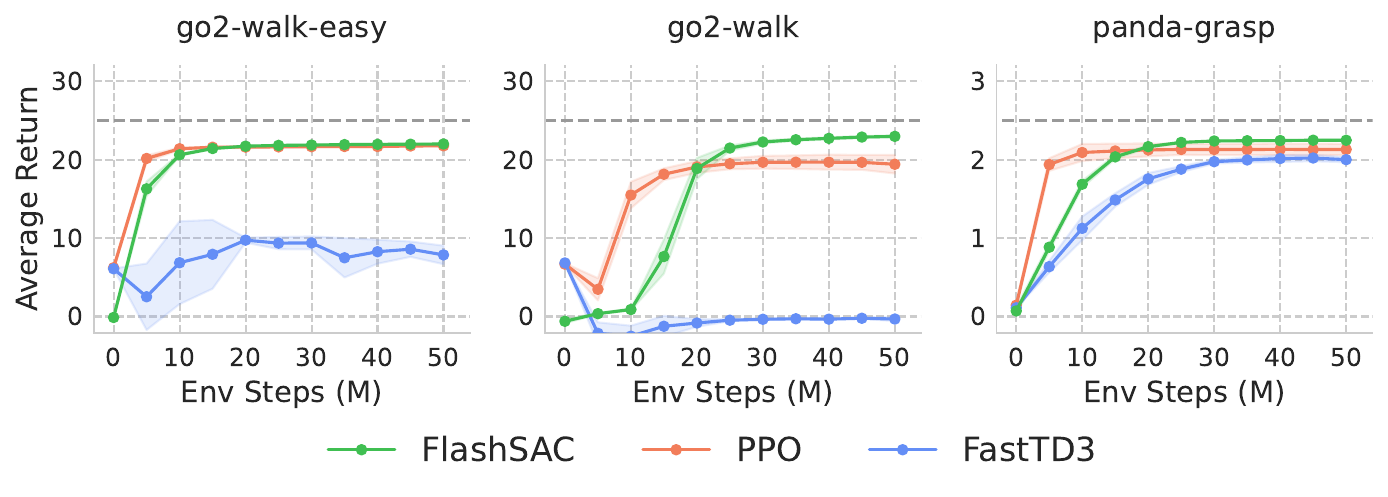}
        \caption{\textbfs{Genesis Learning Curves (Sample Efficiency).} Average episode returns in Genesis environments, plotted against environment steps. Results are averaged over random seeds of each algorithm, with shaded regions indicating $95\%$ bootstrap confidence intervals and dotted lines denoting normalize score.}
        \label{figure:full_result_genesis}
    \end{minipage}
\end{figure}

\clearpage

\subsection{Mujoco (State-based RL, CPU Simulator)}
\label{appendix:full_results_mujoco}

\begin{figure}[h!]
    \centering
    \vspace{0.5em}
    
    \begin{minipage}{\textwidth}
        \centering
        \includegraphics[width=0.99\textwidth]{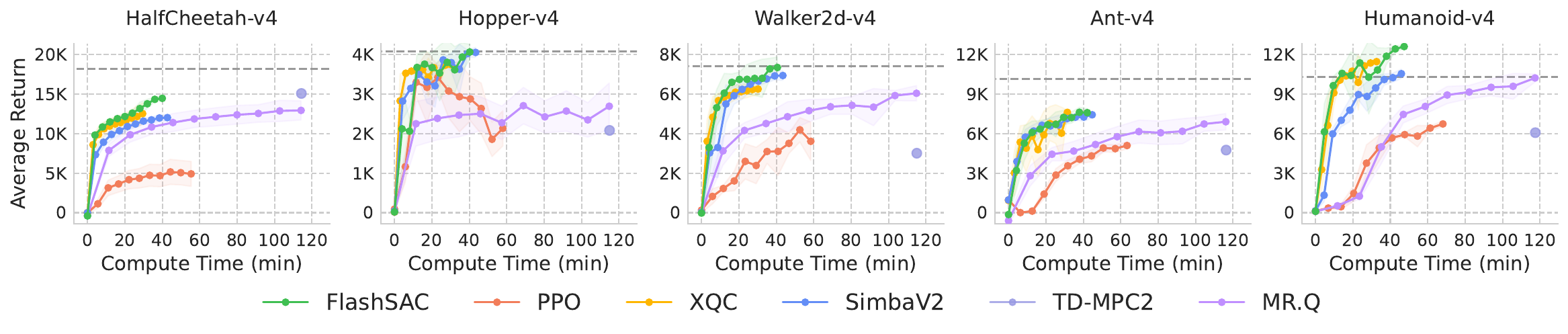}
        \caption{\textbfs{MuJoCo Learning Curves (Compute Efficiency).} Average episode returns in MuJoCo environments, plotted against total compute time.  Results are averaged over random seeds of each algorithm, with shaded regions indicating $95\%$ bootstrap confidence intervals and dotted lines denoting normalized score. All methods are trained for 1M environment steps, except for PPO (4M).}
    \end{minipage}
    
    \vspace{3em}
    
    \begin{minipage}{\textwidth}
        \centering
        \includegraphics[width=0.99\textwidth]{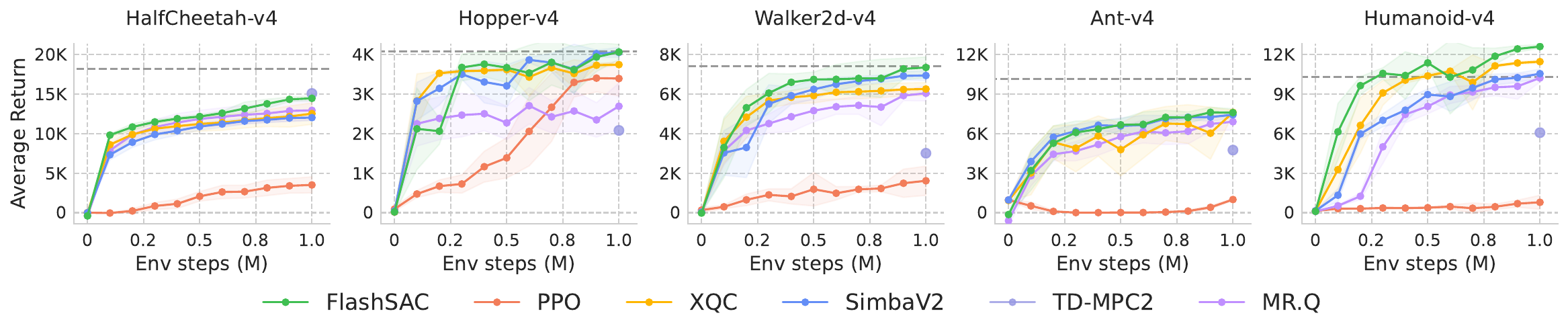}
        \caption{\textbfs{MuJoCo Learning Curves (Sample Efficiency).} Average episode returns in MuJoCo environments, plotted against environment steps. Results are averaged over random seeds of each algorithm, with shaded regions indicating $95\%$ bootstrap confidence intervals and dotted lines denoting normalize score.}
        \label{figure:full_result_mujoco}
    \end{minipage}
\end{figure}

\clearpage

\subsection{DeepMind Control Suite (State-based RL, CPU Simulator)}
\label{appendix:full_results_dmc_state}

\begin{figure}[h!]
    \centering
    \vspace{0.5em}
    
    \begin{minipage}{\textwidth}
        \centering
        \includegraphics[width=0.99\textwidth]{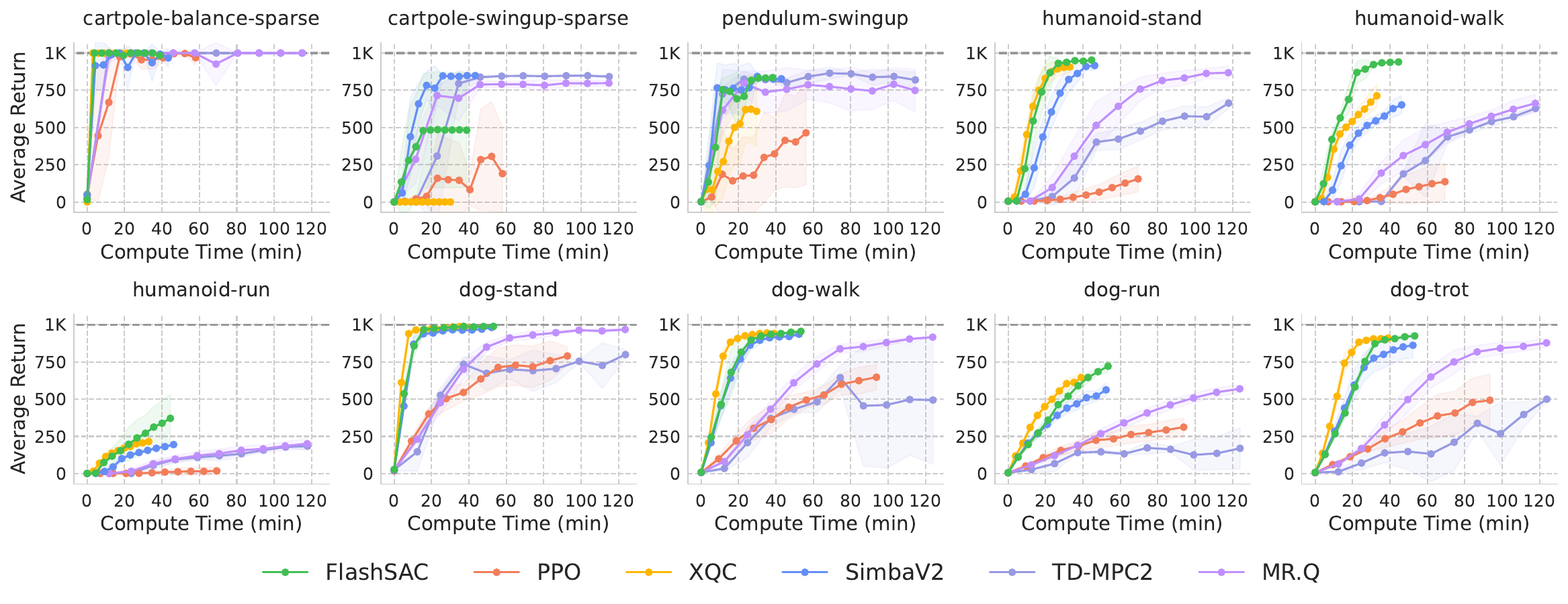}
        \caption{\textbfs{DMC Learning Curves (Compute Efficiency).} Average episode returns in DMC environments, plotted against total compute time.  Results are averaged over random seeds of each algorithm, with shaded regions indicating $95\%$ bootstrap confidence intervals and dotted lines denoting normalized score. All methods are trained for 1M environment steps, except for PPO (4M).}
    \end{minipage}
    
    \vspace{3em}
    
    \begin{minipage}{\textwidth}
        \centering
        \includegraphics[width=0.99\textwidth]{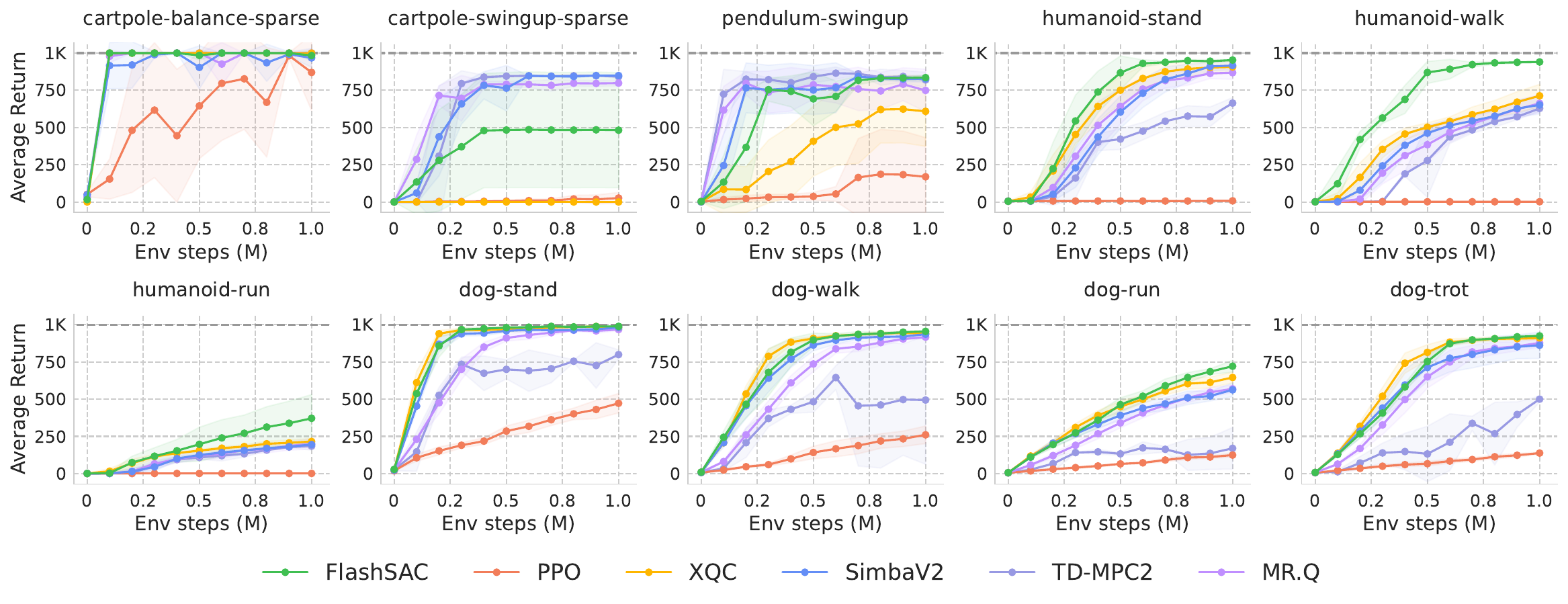}
        \caption{\textbfs{DMC Learning Curves (Sample Efficiency).} Average episode returns in DMC environments, plotted against environment steps. Results are averaged over random seeds of each algorithm, with shaded regions indicating $95\%$ bootstrap confidence intervals and dotted lines denoting normalized score.}
        \label{figure:full_result_dmc_state}
    \end{minipage}
\end{figure}

\clearpage

\subsection{Humanoid Bench (State-based RL, CPU Simulator)}
\label{appendix:full_results_hbench}

\begin{figure}[h!]
    \centering
    
    \begin{minipage}{\textwidth}
        \centering
        \includegraphics[width=0.95\textwidth]{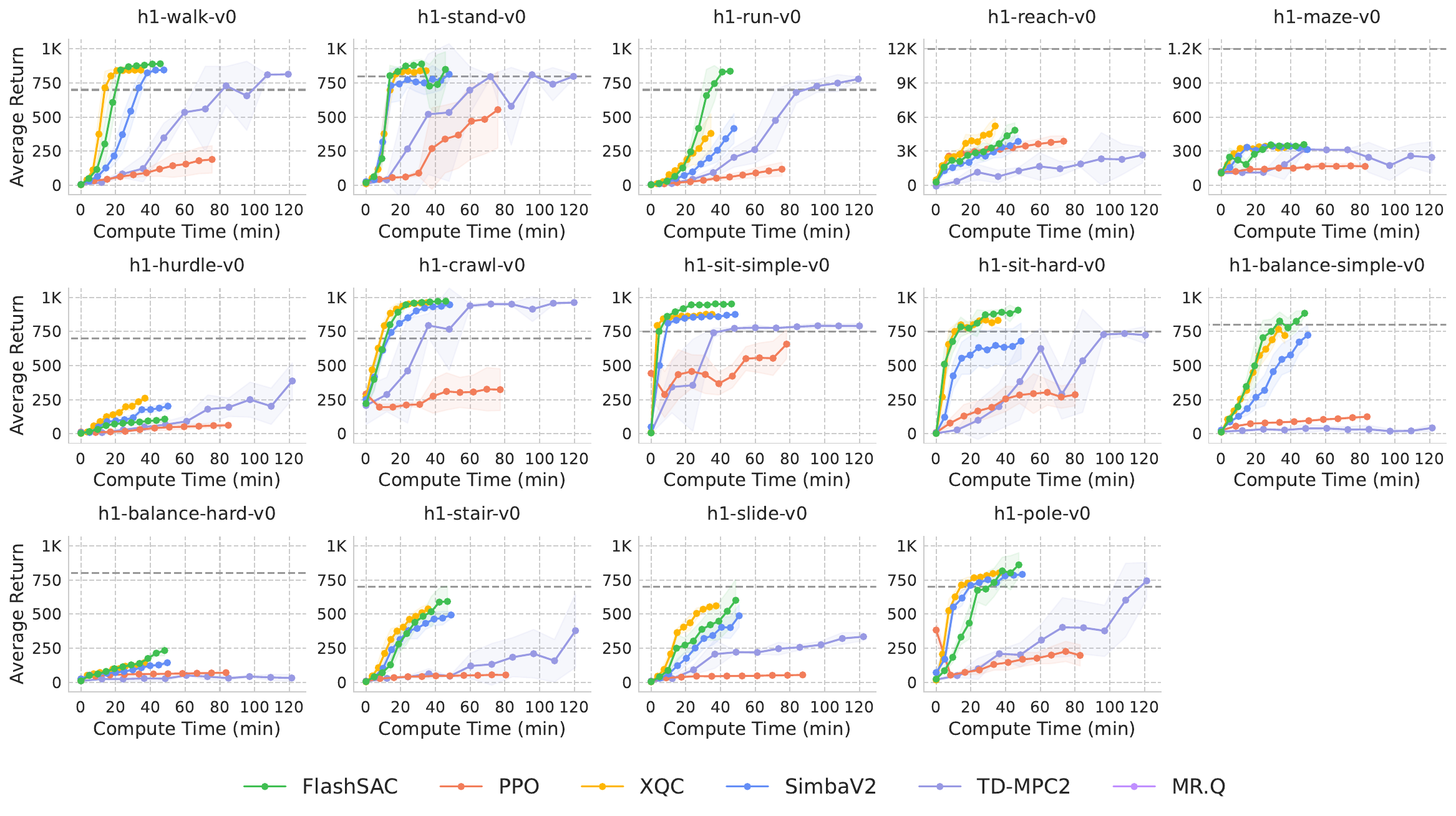}
        \caption{\textbfs{Humanoid Bench Learning Curves (Compute Efficiency).} Average episode returns plotted against total compute time. Results are averaged over random seeds of each algorithm, with shaded regions indicating $95\%$ bootstrap confidence intervals and dotted lines denoting normalized score. All methods are trained for 1M environment steps, except for PPO (4M).}
    \end{minipage}
    
    \vspace{0.5em}
    
    \begin{minipage}{\textwidth}
        \centering
        \includegraphics[width=0.95\textwidth]{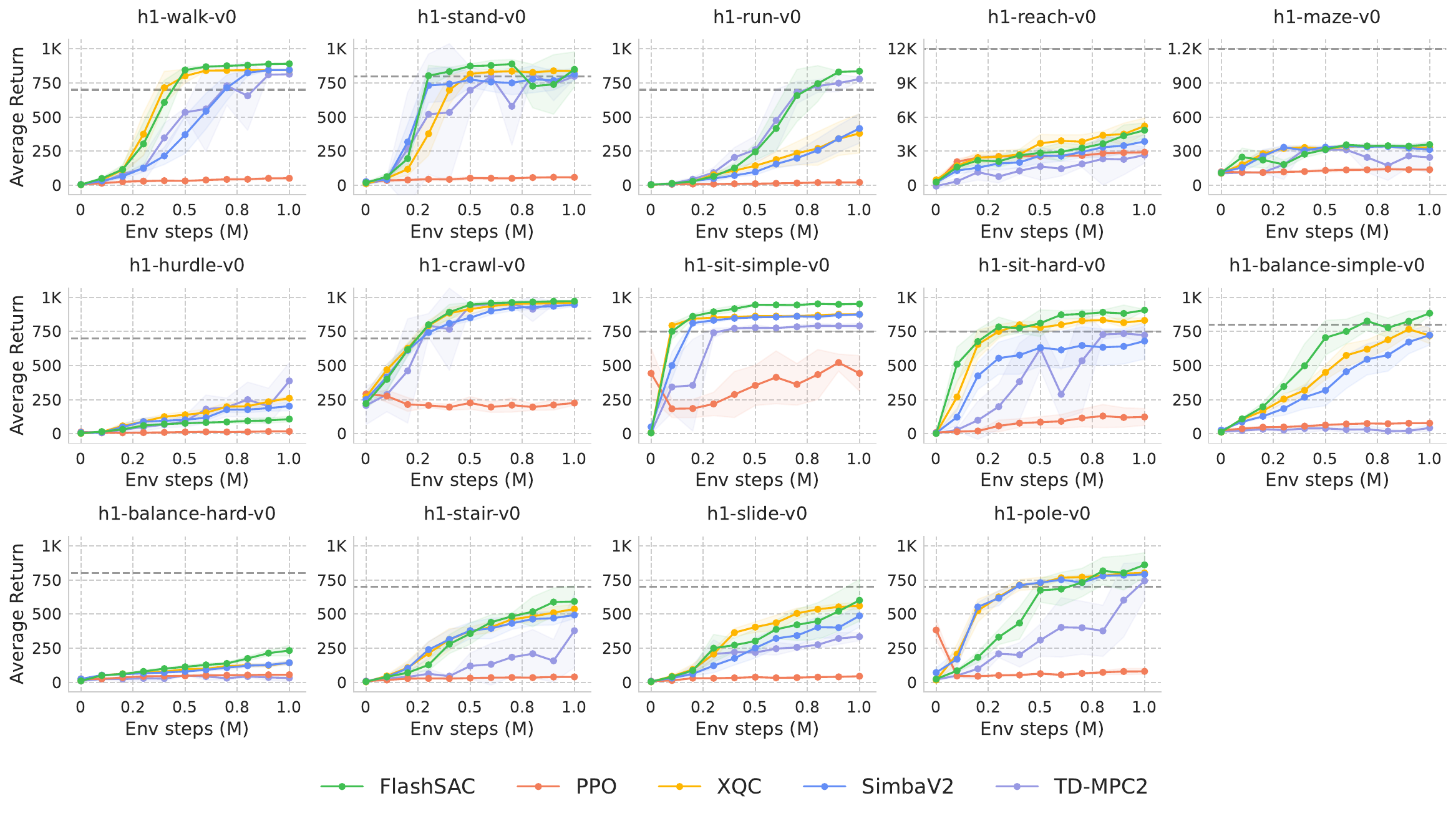}
        \caption{\textbfs{Humanoid Bench Learning Curves (Sample Efficiency).} Average episode returns in Humanoid Bench environments, plotted against environment steps. Results are averaged over random seeds of each algorithm, with shaded regions indicating $95\%$ bootstrap confidence intervals and dotted lines denoting normalize score.}
        \label{figure:full_result_humanoid_bench}
    \end{minipage}
\end{figure}

\clearpage

\subsection{MyoSuite (State-based RL, CPU Simulator)}
\label{appendix:full_results_myosuite}

\begin{figure}[h!]
    \centering
    \vspace{0.5em}
    
    \begin{minipage}{\textwidth}
        \centering
        \includegraphics[width=0.99\textwidth]{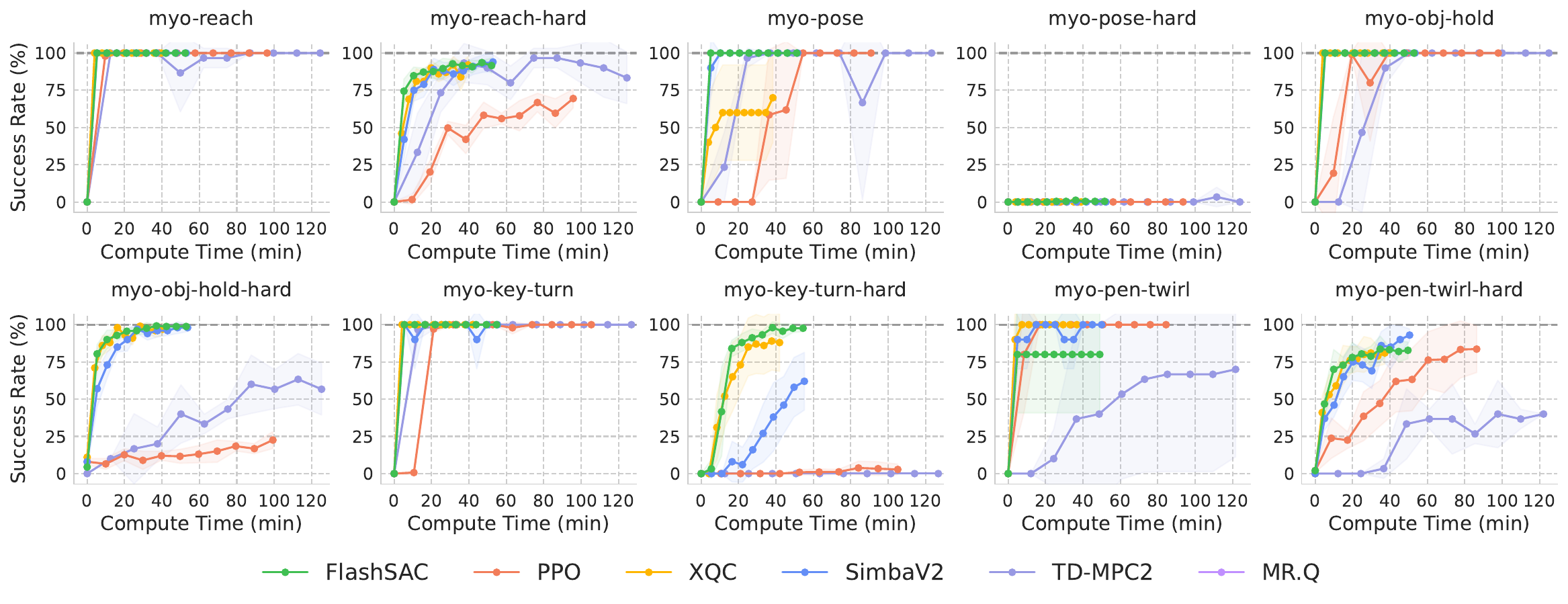}
        \caption{\textbfs{MyoSuite Learning Curves (Compute Efficiency).} Average episode returns in MyoSuite environments, plotted against total compute time.  Results are averaged over random seeds of each algorithm, with shaded regions indicating $95\%$ bootstrap confidence intervals and dotted lines denoting normalized score. All methods are trained for 1M environment steps, except for PPO (4M).}
    \end{minipage}
    
    \vspace{3em}
    
    \begin{minipage}{\textwidth}
        \centering
        \includegraphics[width=0.99\textwidth]{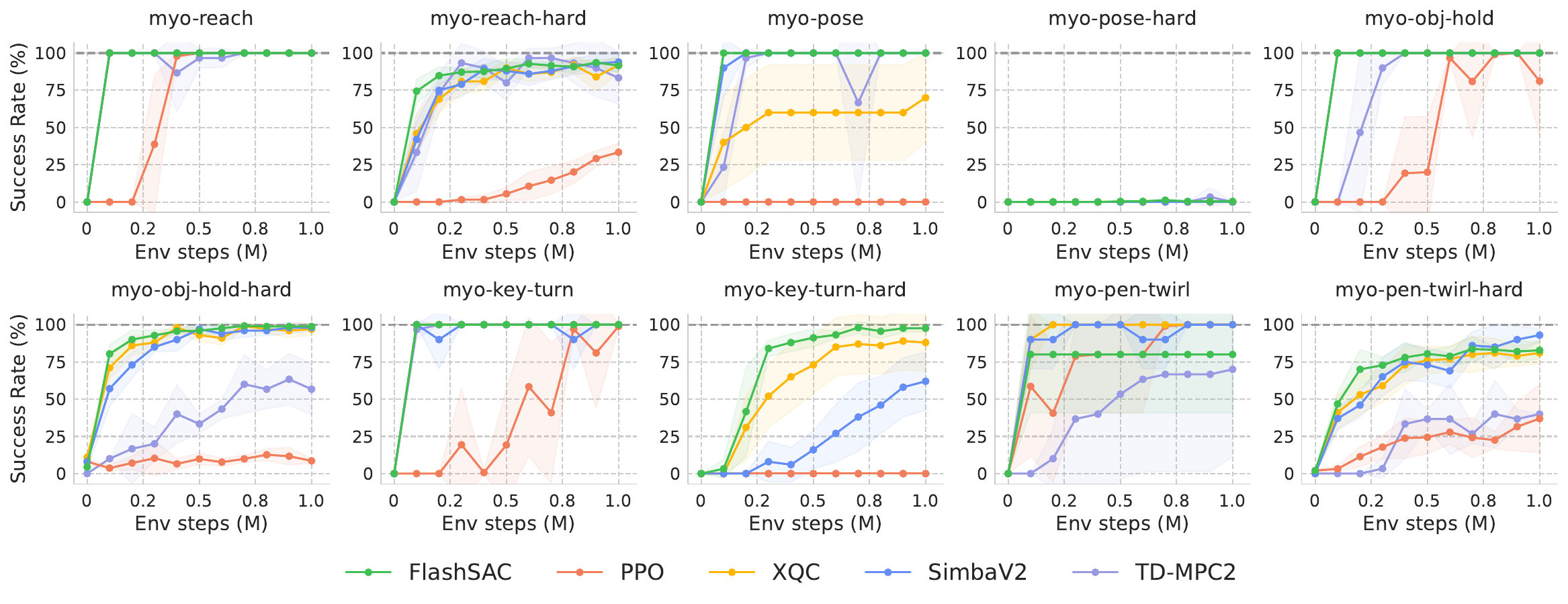}
        \caption{\textbfs{MyoSuite Learning Curves (Sample Efficiency).} Average episode returns in MyoSuite environments, plotted against environment steps. Results are averaged over random seeds of each algorithm, with shaded regions indicating $95\%$ bootstrap confidence intervals and dotted lines denoting normalize score.}
        \label{figure:full_result_myosuite}
    \end{minipage}
\end{figure}

\clearpage

\subsection{Vision-based RL}
\label{appendix:full_results_vision}

\begin{figure}[h!]
    \centering
    \vspace{0.5em}
    
    \begin{minipage}{\textwidth}
        \centering
        \includegraphics[width=0.99\textwidth]{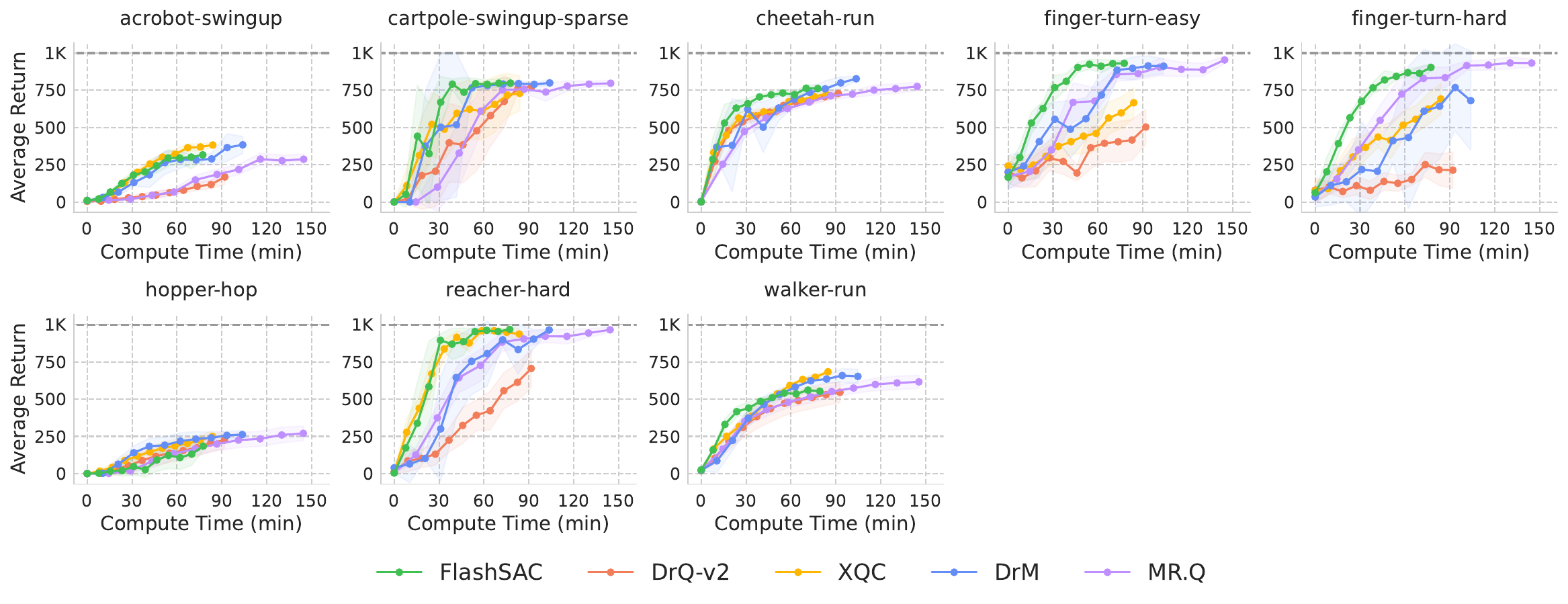}
        \caption{\textbfs{DMC-Visual Learning Curves (Compute Efficiency).} Average episode returns in DMC-Visual environments, plotted against total compute time.  Results are averaged over random seeds of each algorithm, with shaded regions indicating $95\%$ bootstrap confidence intervals and dotted lines denoting normalized score. All methods are trained for 1M environment steps.}
    \end{minipage}
    
    \vspace{3em}
    
    \begin{minipage}{\textwidth}
        \centering
        \includegraphics[width=0.99\textwidth]{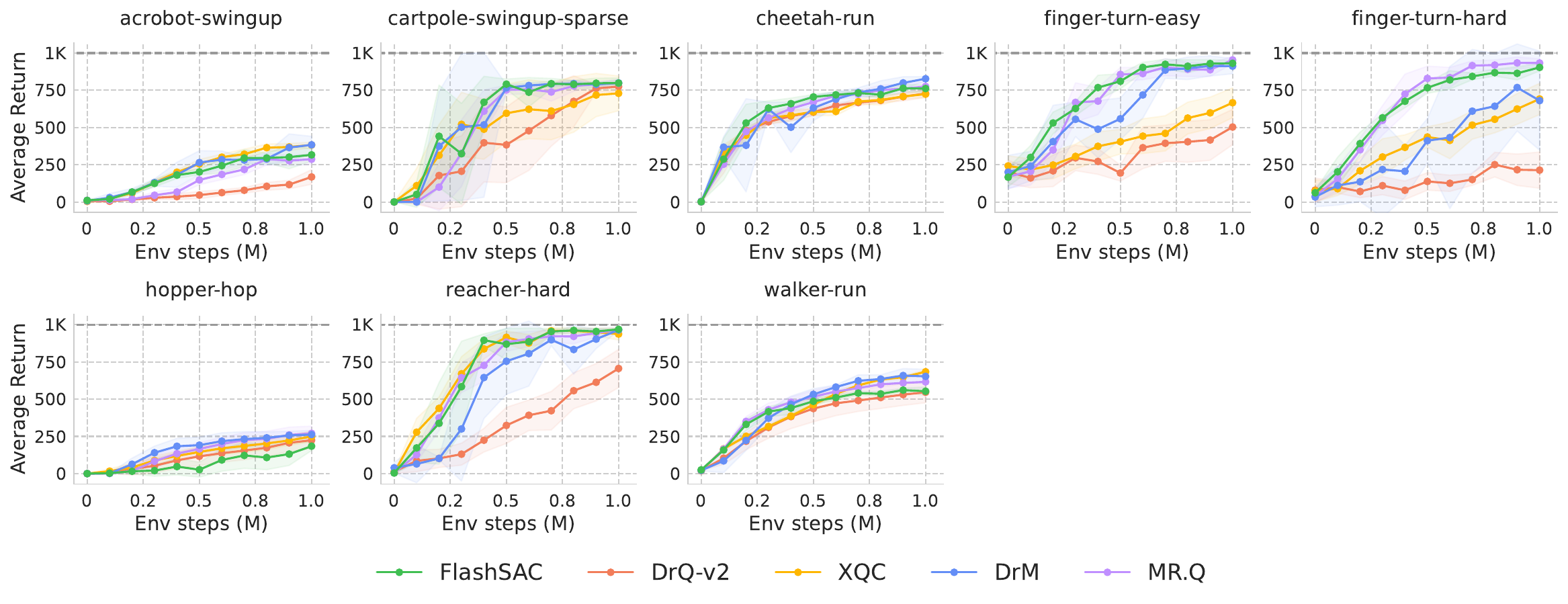}
        \caption{\textbfs{DMC-Visual Learning Curves (Sample Efficiency).} Average episode returns in DMC-Visual environments, plotted against environment steps. Results are averaged over random seeds of each algorithm, with shaded regions indicating $95\%$ bootstrap confidence intervals and dotted lines denoting normalized score.}
        \label{figure:full_result_dmc_visual}
    \end{minipage}
\end{figure}

\clearpage

\subsection{Ablation Study}
\label{appendix:full_results_ablation_component}

\begin{figure}[h!]
    \centering
    \vspace{0.1em}
    
    \begin{minipage}{\textwidth}
        \centering
        \includegraphics[width=0.85\textwidth]{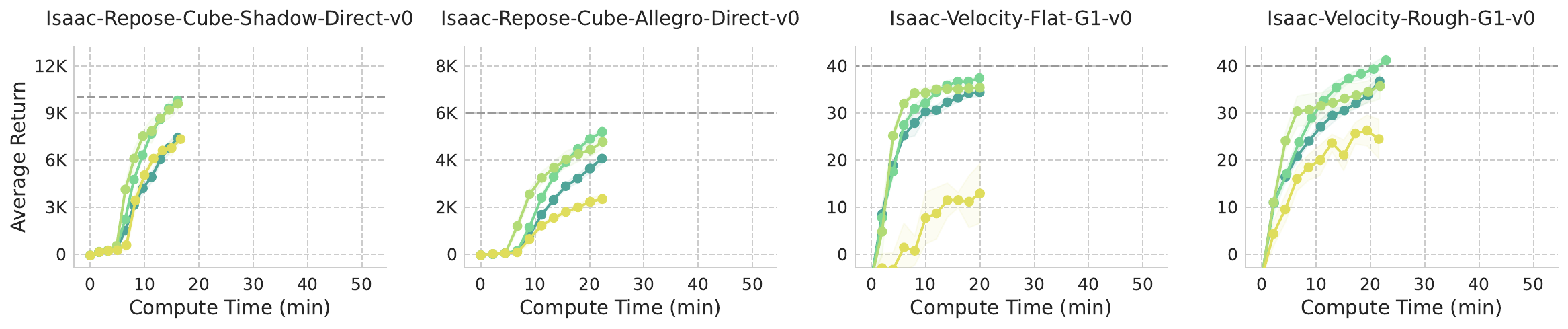}
    \end{minipage}
    
    \vspace{0.1em}
    
    \begin{minipage}{\textwidth}
        \centering
        \includegraphics[width=0.85\textwidth]{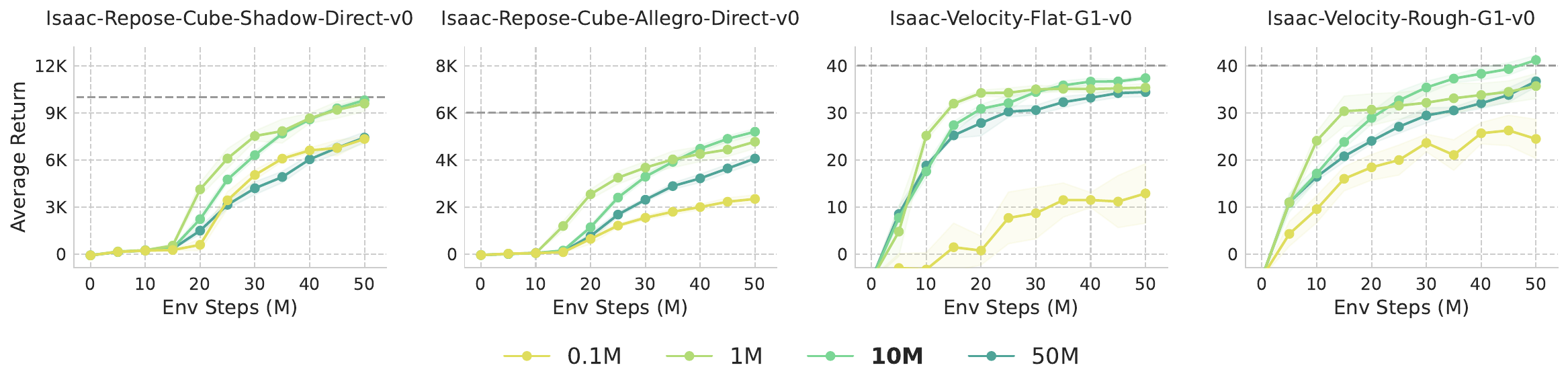}
        \vspace{-2mm}
        \captionof{figure}{\textbfs{Ablation:  Buffer Size.} Each configuration indicates the maximum size of the replay buffer. Results are averaged over random seeds of each configuration, with shaded regions indicating $95\%$ bootstrap confidence intervals and dotted lines denoting normalized score.}
        \label{fig:hyperparam_buffer_sample}
    \end{minipage}
\end{figure}

\begin{figure}[h!]
    \centering
    \vspace{0.1em}
    
    \begin{minipage}{\textwidth}
        \centering
        \includegraphics[width=0.85\textwidth]{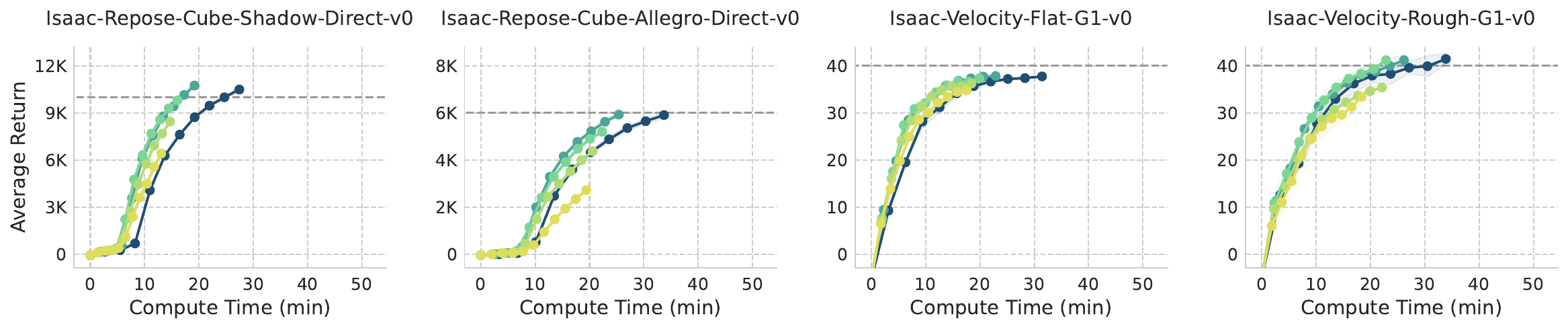}
    \end{minipage}
    
    \vspace{0.1em}
    
    \begin{minipage}{\textwidth}
        \centering
        \includegraphics[width=0.85\textwidth]{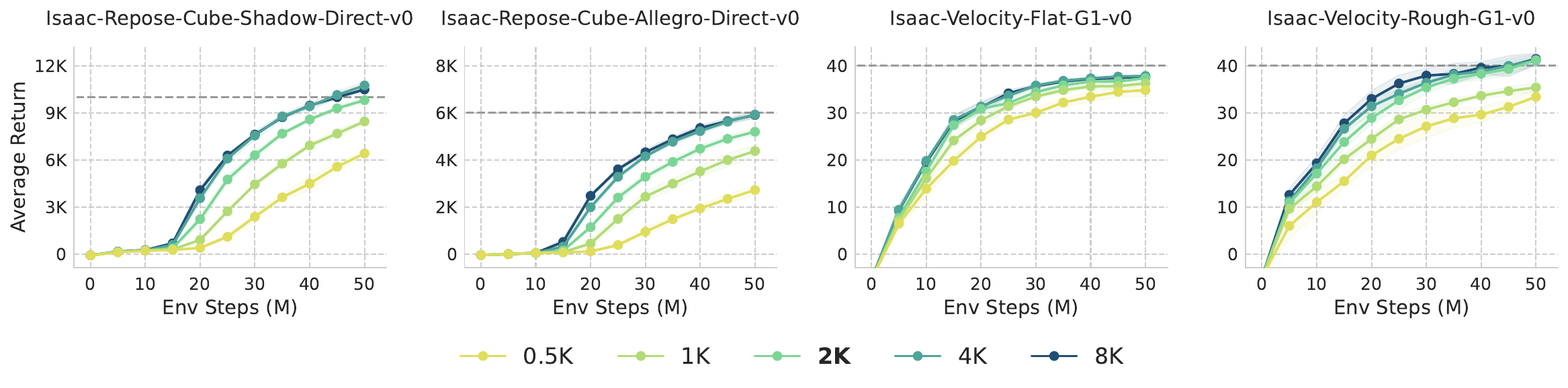}
        \vspace{-2mm}
        \captionof{figure}{\textbfs{Ablation: Batch Size.} Each configuration indicates the size of the mini-batch used for training. Results are averaged over random seeds of each configuration, with shaded regions indicating $95\%$ bootstrap confidence intervals and dotted lines denoting normalized score.}
        \label{fig:hyperparam_batch_sample}
    \end{minipage}
\end{figure}

\clearpage

\begin{figure}[h!]
    \centering    
    \begin{minipage}{\textwidth}
        \centering
        \includegraphics[width=0.85\textwidth]{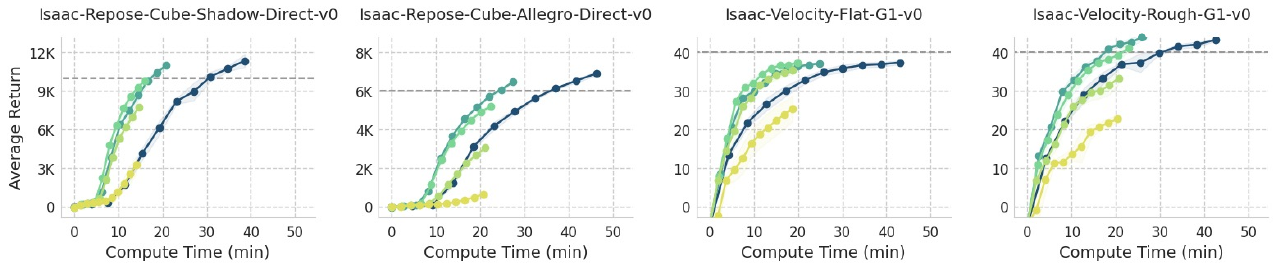}
    \end{minipage}
    
    \vspace{0.1em}
    
    \begin{minipage}{\textwidth}
        \centering
        \includegraphics[width=0.85\textwidth]{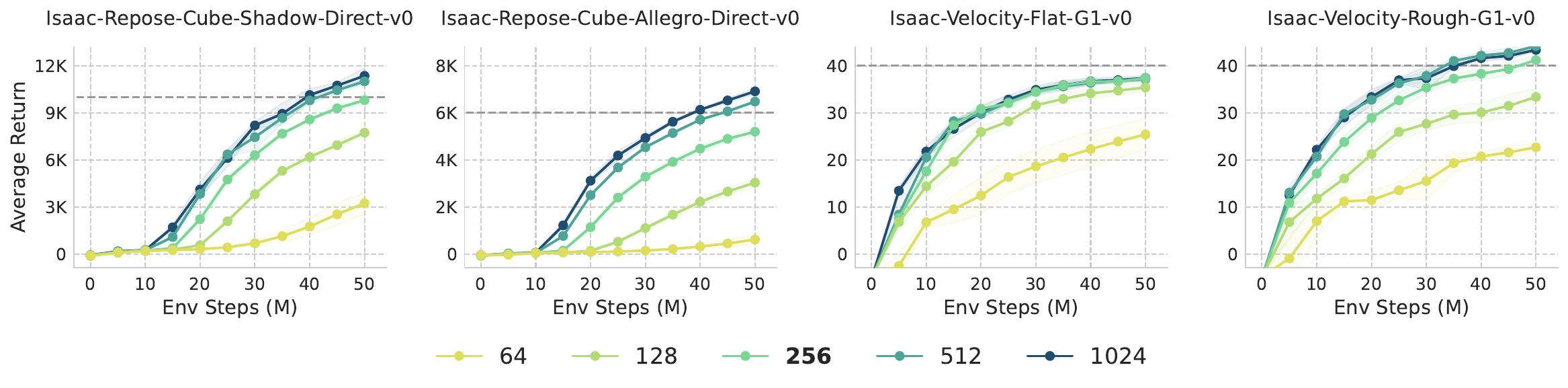}
        \vspace{-2mm}
        \captionof{figure}{\textbfs{Ablation: Network Width.} Each configuration indicates the hidden dimension of the critic network, while the actor network is proportionally scaled as well. Results are averaged over random seeds of each configuration, with shaded regions indicating $95\%$ bootstrap confidence intervals and dotted lines denoting normalized score.}
        \label{fig:hyperparam_width_sample}
    \end{minipage}
\end{figure}

\begin{figure}[h!]
    \centering    
    \begin{minipage}{\textwidth}
        \centering
        \includegraphics[width=0.85\textwidth]{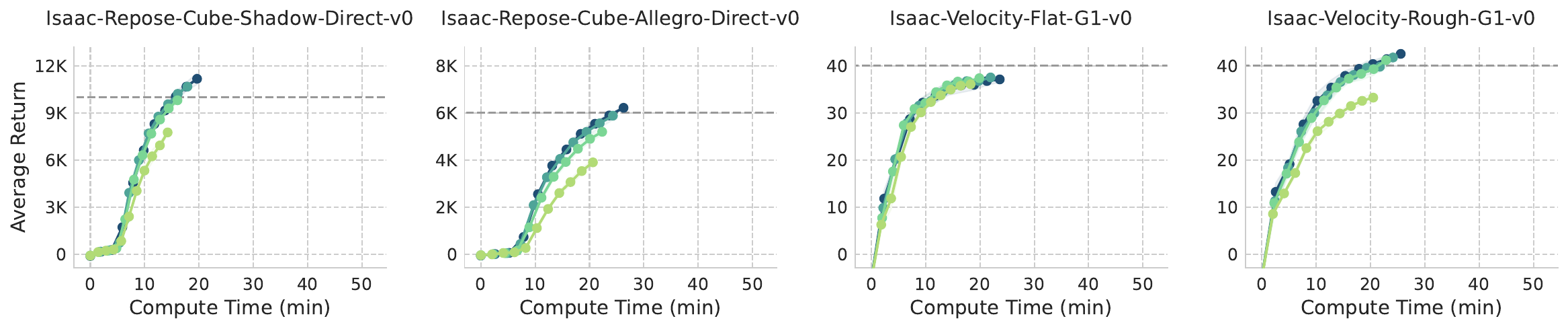}
    \end{minipage}
    
    \vspace{0.1em}
    
    \begin{minipage}{\textwidth}
        \centering
        \includegraphics[width=0.85\textwidth]{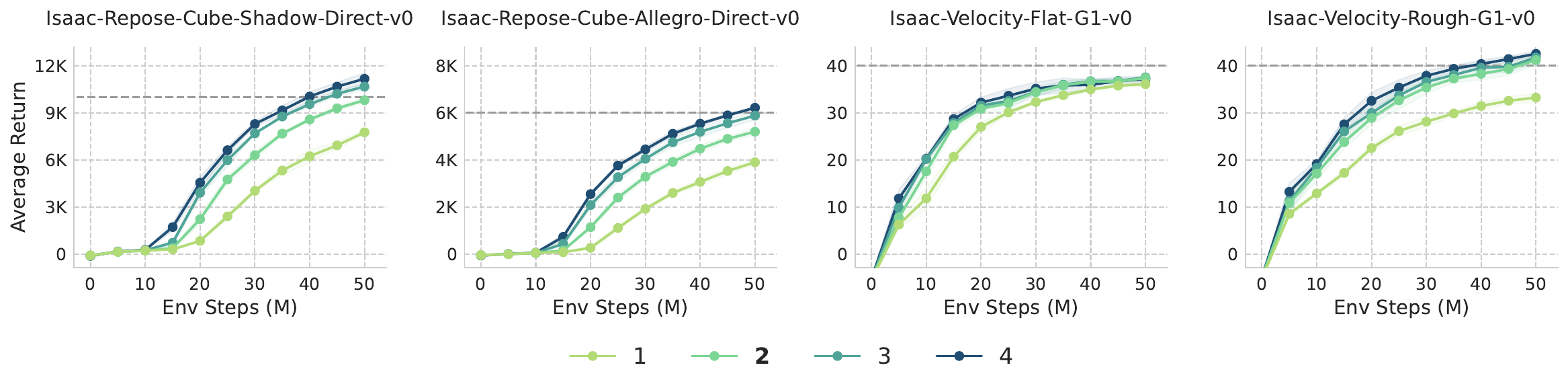}
        \vspace{-2mm}
        \captionof{figure}{\textbfs{Ablation: Network Depth.} Each configuration indicates the number of residual blocks in both actor and critic networks. Average episode returns in IsaacLab environments, plotted against the number of environment steps. Results are averaged over random seeds of each configuration, with shaded regions indicating $95\%$ bootstrap confidence intervals and dotted lines denoting normalized score. }
        \label{fig:hyperparam_depth_sample}
    \end{minipage}
\end{figure}

\clearpage

\begin{figure}[h!]
    \centering
    \vspace{0.1em}
    
    \begin{minipage}{\textwidth}
        \centering
        \includegraphics[width=0.85\textwidth]{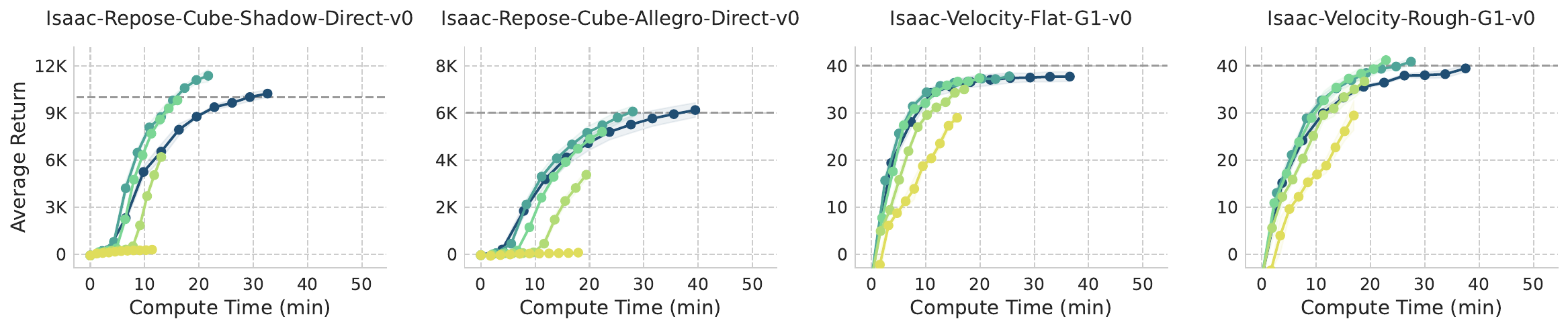}
    \end{minipage}
    
    \vspace{0.1em}
    
    \begin{minipage}{\textwidth}
        \centering
        \includegraphics[width=0.85\textwidth]{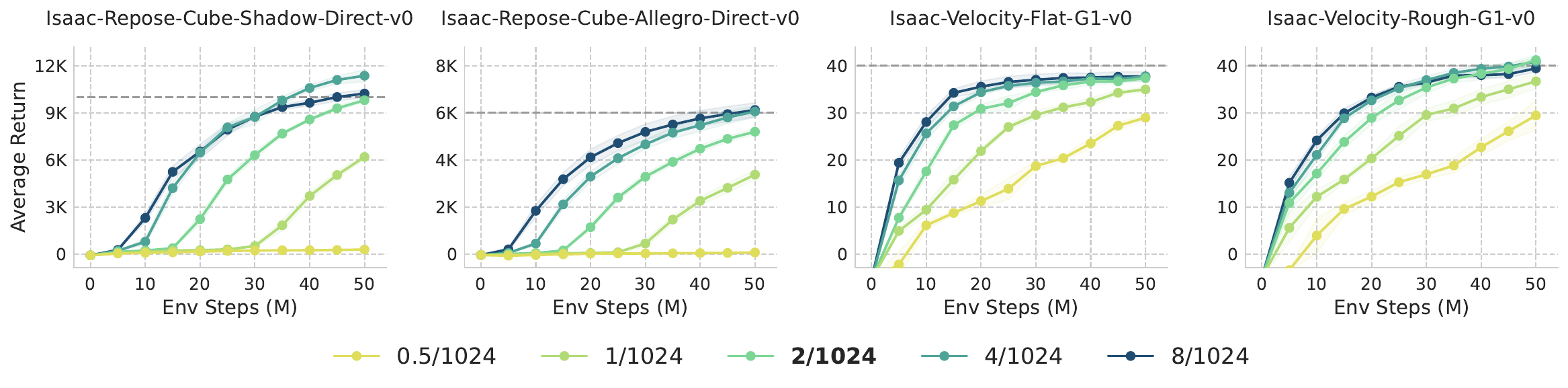}
        \vspace{-2mm}
        \captionof{figure}{\textbfs{Ablation: UTD Ratio.} Each configuration indicates the update-to-data ratio for training. Average episode returns in IsaacLab environments, plotted against environment steps. Results are averaged over random seeds of each configuration, with shaded regions indicating $95\%$ bootstrap confidence intervals and dotted lines denoting normalized score.}
        \label{fig:hyperparam_batch_sample}
    \end{minipage}
\end{figure}

\begin{figure}[h!]
    \centering
    \vspace{0.5em}
    
    \begin{minipage}{\textwidth}
        \centering
        \includegraphics[width=0.85\textwidth]{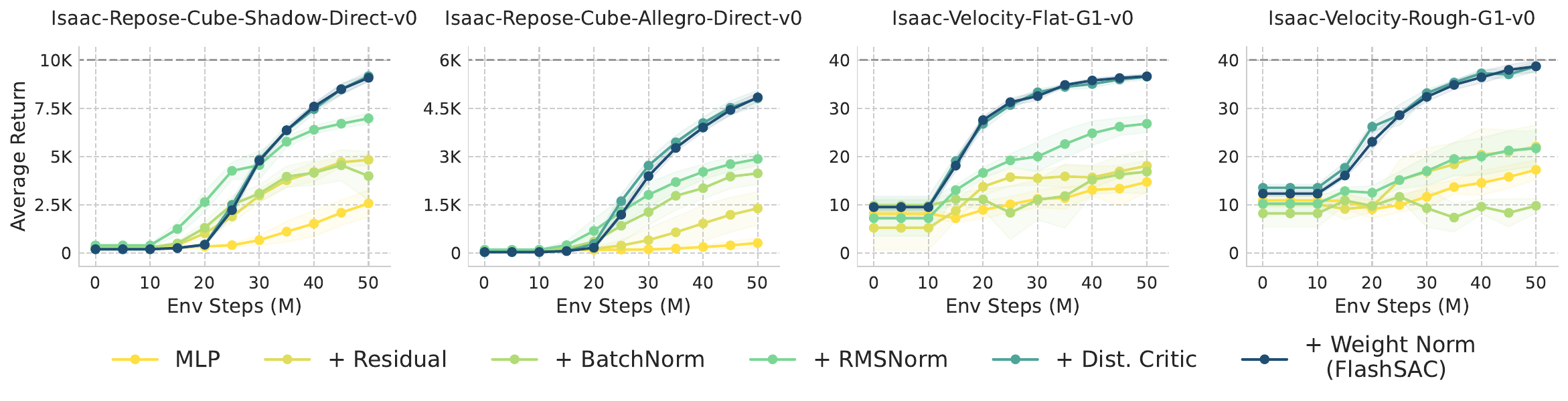}
        \vspace{-2mm}
        \captionof{figure}{\textbfs{Architectural Ablation Learning Curves.} Average episode returns in IsaacLab environments, plotted against environment steps. Results are averaged over random seeds of each configuration, with shaded regions indicating $95\%$ bootstrap confidence intervals and dotted lines denoting normalized score.}
    \end{minipage}

    \vspace{3em}
\end{figure}

\clearpage

\begin{figure}[h!]
    \centering
    \vspace{0.5em}
    
    \begin{minipage}{\textwidth}
        \centering
        \includegraphics[width=0.85\textwidth]{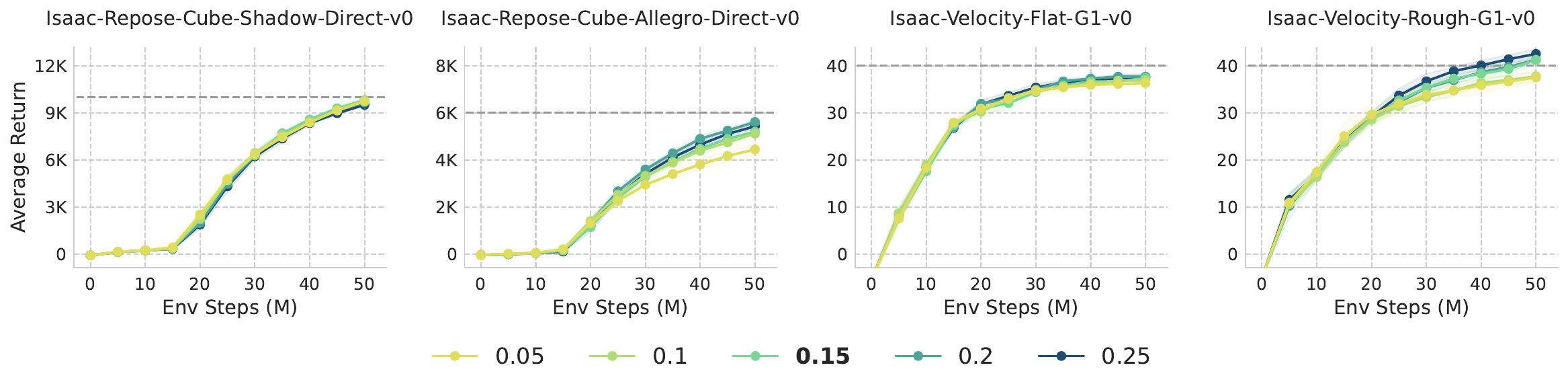}
        \vspace{-2mm}
        \captionof{figure}{\textbfs{Ablation: Entropy target $\sigma_{tgt}$.} Each configuration indicates the target standard deviation used for automatic temperature tuning. Average episode returns in IsaacLab environments, plotted against environment steps. Results are averaged over random seeds of each configuration, with shaded regions indicating 95\% bootstrap confidence intervals and dotted lines denoting normalized score. 
        }
    \end{minipage}
\end{figure}

\begin{figure}[h!]
    \centering
    \vspace{0.5em}
    \begin{minipage}{\textwidth}
        \centering
        \includegraphics[width=0.85\textwidth]{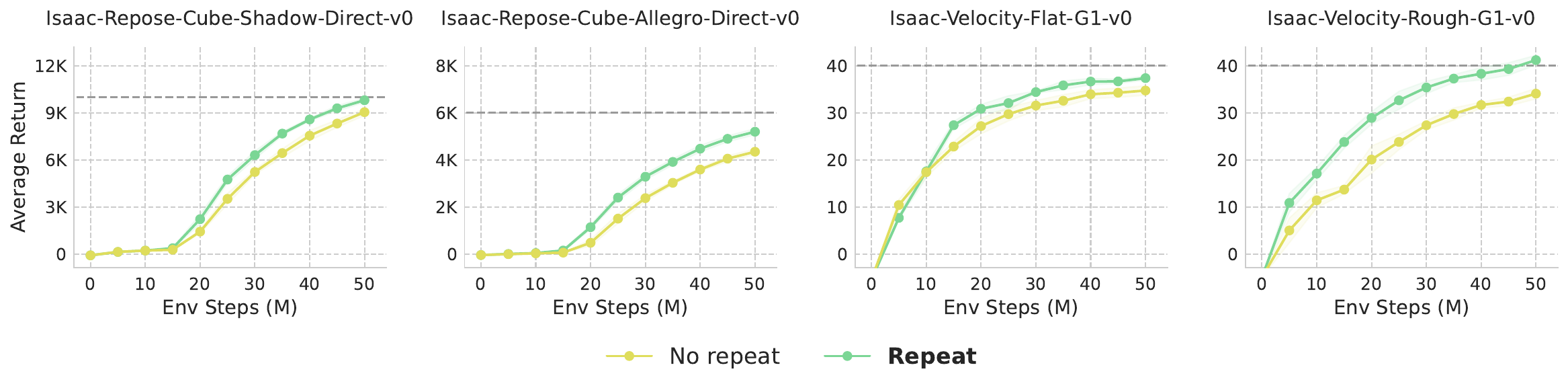}
        \vspace{-2mm}
        \captionof{figure}{\textbfs{Ablation: Noise Repeat.} Each configuration indicates whether exploration via noise repeat is enabled or not. Average episode returns in IsaacLab environments, plotted against environment steps. Results are averaged over random seeds of each configuration, with shaded regions indicating 95\% bootstrap confidence intervals and dotted lines denoting normalized score. 
        }
    \end{minipage}

    \vspace{3em}
\end{figure}

\end{document}

%% file: tables/info_isaaclab.tex
\parbox{0.95\textwidth}{
\captionof{table}{\textbfs{IsaacLab environments.} We evaluate 12 tasks from IsaacLab spanning gripper manipulation, dexterous manipulation, quadruped locomotion, and humanoid locomotion. Normalized scores are defined per environment and correspond to near-asymptotic performance achieved after extended training.}
\label{tab:isaaclab_environment}
\centering
    \small
    \begin{tabular}{lcccc}
        \toprule
        \textbf{Task} & \textbf{Observation dim} & \textbf{Action dim} & \textbf{Normalize Score} \\
        \midrule
        \texttt{Isaac-Repose-Cube-Shadow-Direct-v0}  & 157 & 20 & 10000 \\
        \texttt{Isaac-Repose-Cube-Allegro-Direct-v0} & 124 & 16 & 6000 \\
        \texttt{Isaac-Velocity-Flat-G1-v0} & 123 & 37 & 40 \\
        \texttt{Isaac-Velocity-Rough-G1-v0} & 310 & 37 & 40 \\
        \texttt{Isaac-Velocity-Flat-H1-v0}  & 69 & 19 & 40 \\
        \texttt{Isaac-Velocity-Rough-H1-v0}& 256 & 19 & 40 \\
        \texttt{Isaac-Lift-Cube-Franka-v0} & 36 & 8 & 160 \\
        \texttt{Isaac-Open-Drawer-Franka-v0}  & 31 & 8 & 100 \\
        \texttt{Isaac-Velocity-Flat-Anymal-C-v0} & 48 & 12 & 30 \\
        \texttt{Isaac-Velocity-Rough-Anymal-C-v0}  & 235 & 12 & 30 \\
        \texttt{Isaac-Velocity-Flat-Anymal-D-v0}  & 48 & 12 & 30 \\
        \texttt{Isaac-Velocity-Rough-Anymal-D-v0}& 235 & 12 & 30 \\
        \bottomrule
    \end{tabular}
}

%% file: tables/info_mjp.tex
\parbox{0.95\textwidth}{
\captionof{table}{\textbfs{MuJoCo Playground environments.} We evaluate four humanoid locomotion tasks from MuJoCo Playground. Normalized scores are scaled to 40, corresponding to asymptotic performance achieved after extended training.  If the environment supports asymmetric observation, the privileged observation size is given in parentheses.}
\label{tab:mjp_environment}
\centering
    \small
    \begin{tabular}{lccccc}
        \toprule
        \textbf{Task} & \textbf{Observation dim} & \textbf{Action dim} & \textbf{Normalize Score} \\
        \midrule
        \texttt{G1JoystickRoughTerrain} & 103 (216) & 29 & 40 \\
        \texttt{G1JoystickFlatTerrain} & 103 (216) & 29 & 40 \\
        \texttt{T1JoystickRoughTerrain} & 85 (180) & 23 & 40 \\
        \texttt{T1JoystickFlatTerrain} & 85 (180) & 23 & 40 \\
        \bottomrule
    \end{tabular}
}

%% file: tables/info_maniskill.tex
\parbox{0.95\textwidth}{
\captionof{table}{\textbfs{ManiSkill Environments.} We evaluate 6 ManiSkill gripper-based manipulation environments. Normalized scores correspond to the maximum success rate, where 100 denotes a 100\% success rate.}
\label{tab:maniskill_environment}
\centering
    \small
    \begin{tabular}{lcccc}
        \toprule
        \textbf{Task}& \textbf{Observation dim} & \textbf{Action dim} & \textbf{Normalize Score} \\
        \midrule
        \texttt{PickSingleYCB-v1}  & 45 & 8 & 100 \\
        \texttt{PegInsertionSide-v1}  & 43 & 8 & 100 \\
        \texttt{LiftPegUpright-v1} & 32 & 8 & 100 \\
        \texttt{PokeCube-v1} & 54 & 8 & 100 \\
        \texttt{PullCube-v1}  & 35 & 8 & 100 \\
        \texttt{RollBall-v1}  & 44 & 8 & 100 \\
        \bottomrule
    \end{tabular}
}

%% file: tables/info_genesis.tex
\parbox{0.95\textwidth}{
\captionof{table}{\textbfs{Genesis environments.} We evaluate 3 reinforcement learning tasks from Genesis. Normalized scores are defined per environment and correspond to near-asymptotic performance achieved after extended training. If the environment supports asymmetric observation, the privileged observation size is given in parentheses.}
\label{tab:genesis_environment}
\centering
    \small
    \begin{tabular}{lccccc}
        \toprule
        \textbf{Task}  & \textbf{Observation dim} & \textbf{Action dim} & \textbf{Normalize Score} \\
        \midrule
        \texttt{go2-walk\_easy}  & 45 & 12 & 25 \\
        \texttt{go2-walk} & 45 (60) & 12 & 25 \\
        \texttt{panda-grasp}  & 14 & 6 & 2.5 \\
        \bottomrule
    \end{tabular}
}

%% file: tables/info_mujoco.tex
\parbox{0.95\textwidth}{
\captionof{table}{\textbfs{MuJoCo Environments.} We evaluate five standard MuJoCo environments. Normalized scores are computed relative to the performance of TD7 \cite{fujimoto2023td7} after 5M training steps.}
\label{tab:mujoco_environment}
\centering
\small
\begin{tabular}{lccccc}
    \toprule
    \textbf{Task} & \textbf{Observation dim} & \textbf{Action dim} & \textbf{Random Score} & \textbf{Normalize Score} \\
    \midrule
    \texttt{HalfCheetah-v4} & 17 & 6 & -289.415 & 18165 \\
    \texttt{Hopper-v4} & 11 & 3 & 18.791 & 4075 \\
    \texttt{Walker2d-v4} & 17 & 6 & 2.791 & 7397 \\
    \texttt{Ant-v4} & 27 & 8 & -70.288 & 10133 \\
    \texttt{Humanoid-v4} & 376 & 17 & 120.423 & 10281 \\
    \bottomrule
\end{tabular}}

%% file: tables/info_dmc.tex
\parbox{0.95\textwidth}{
\captionof{table}{\textbfs{DMC Environments.} We evaluate 10 DeepMind Control Suite tasks, including humanoid and dog embodiments and sparse-reward settings. Normalized scores correspond to the theoretical maximum.}
\label{tab:dmc_environment}
\centering
    \small
    \begin{tabular}{lcccc}
        \toprule
        \textbf{Task} & \textbf{Observation dim} & \textbf{Action dim} & \textbf{Normalize Score} \\
        \midrule
        \texttt{cartpole-balance\_sparse} & 5 & 1 & 1000 \\
        \texttt{cartpole-swingup\_sparse} & 5 & 1 & 1000 \\
        \texttt{pendulum-swingup} & 3 & 1 & 1000 \\
        \texttt{humanoid-stand} & 67 & 21 & 1000 \\
        \texttt{humanoid-walk} & 67 & 21 & 1000 \\
        \texttt{humanoid-run} & 67 & 21 & 1000 \\
        \texttt{dog-stand} & 223 & 38 & 1000 \\
        \texttt{dog-walk} & 223 & 38 & 1000 \\
        \texttt{dog-run} & 223 & 38 & 1000 \\
        \texttt{dog-trot} & 223 & 38 & 1000 \\
        \bottomrule
    \end{tabular}
}

%% file: tables/info_hbench.tex
\parbox{0.95\textwidth}{
\captionof{table}{\textbfs{HumanoidBench Environments.} We evaluate 14 humanoid locomotion tasks without hand control from HumanoidBench. Normalized scores correspond to task success.}
\label{tab:hbench_environment}
\centering
    \small
    \begin{tabular}{lccccc}
        \toprule
        \textbf{Task} & \textbf{Observation dim} & \textbf{Action dim} & \textbf{Random Score} & \textbf{Normalize Score} \\
        \midrule
        \texttt{h1-walk-v0} & 51 & 19 & 2.38 & 700 \\
        \texttt{h1-stand-v0} & 51 & 19 & 10.55 & 800 \\
        \texttt{h1-run-v0} & 51 & 19 & 2.02 & 700 \\
        \texttt{h1-reach-v0} & 57 & 19 & 260.30 & 12000 \\
        \texttt{h1-maze-v0} & 51 & 19 & 106.44 & 1200 \\
        \texttt{h1-hurdle-v0} & 51 & 19 & 2.21 & 700 \\
        \texttt{h1-crawl-v0} & 51 & 19 & 272.66 & 700 \\
        \texttt{h1-sit\_simple-v0} & 51 & 19 & 9.40 & 750 \\
        \texttt{h1-sit\_hard-v0} & 64 & 19 &  2.45 & 750 \\
        \texttt{h1-balance\_simple-v0} & 64 & 19 & 9.40 & 800 \\
        \texttt{h1-balance\_hard-v0} & 77 & 19 & 9.04 & 800 \\
        \texttt{h1-stair-v0} & 51 & 19 & 3.11 & 700 \\
        \texttt{h1-slide-v0} & 51 & 19 & 3.19 & 700 \\
        \texttt{h1-pole-v0} & 51 & 19 & 20.09 & 700 \\
        \bottomrule
    \end{tabular}
}

%% file: tables/info_myosuite.tex
\parbox{0.95\textwidth}{
\captionof{table}{\textbfs{Myosuite Environments.} We evaluate 10 Myosuite environments. Normalized scores correspond to the maximum success rate, where 100 denotes a 100\% success rate.}
\label{tab:myosuite_environment}
\small
\centering
\begin{tabular}{lccccc}
    \toprule
    \textbf{Task} & \textbf{Observation dim} & \textbf{Action dim} & \textbf{Normalize Score} \\
    \midrule
    \texttt{myo-reach} & 115 & 39 & 100 \\
    \texttt{myo-reach-hard} & 115 & 39 & 100 \\
    \texttt{myo-pose} & 108 & 39 & 100 \\
    \texttt{myo-pose-hard} & 108 & 39 & 100 \\
    \texttt{myo-obj-hold} & 91 & 39 & 100 \\
    \texttt{myo-obj-hold-hard} & 91 & 39 & 100 \\
    \texttt{myo-key-turn} & 93 & 39 & 100 \\
    \texttt{myo-key-turn-hard} & 93 & 39 & 100 \\
    \texttt{myo-pen-twirl} & 83 & 39 & 100 \\
    \texttt{myo-pen-twirl-hard} & 83 & 39 & 100 \\
    \bottomrule
\end{tabular}}

%% file: tables/hyperparam_gpu.tex
\parbox{0.95\textwidth}{
\captionof{table}{\textbfs{Hyperparameters (GPU-based Simulators).} \textsc{FlashSAC} hyperparameters used in benchmarks that support massive parallel environments (IsaacLab, MuJoCo Playground, ManiSkill, and Genesis).}
\label{table:hyperparameters_gpusim}
\centering
\small
\begin{tabular}{llll}
\toprule
 & \textbf{Hyperparameter} & \textbf{Notation} & \textbf{Value} \\
 \midrule
\multirow{5}{*}{\textbf{Common}} 
 & Parallel environments & - & 1024 \\
 & Replay buffer capacity & - & 10M \\
 & Batch size & - & 2048 \\
 & Update-to-data (UTD) ratio & - & 2/2048 \\
 & TD steps (n-step) & $n$ & 1 \\
 \midrule
\multirow{3}{*}{\textbf{Actor}}  & Number of blocks & - & 2 \\
 & Hidden dimension & $d_{actor}$ & 128 \\
 & Update delay & - & 2 \\
 \midrule
\multirow{7}{*}{\textbf{Critic}}  & Number of blocks &  & 2 \\
 & Hidden dimension & $d_{critic}$ & 256 \\
 & Target critic momentum & $\tau$ & 0.01 \\
 & Number of critics & - & 2 \\
 & Value prediction type & - & Categorical \\
 & Categorical Support & $[G_{\min}, G_{\max}]$ & {[}-5,5{]} \\
 & Number of bins & $n_{atoms}$ & 101 \\
 \midrule
\multirow{2}{*}{\textbf{Temperature}}  & Entropy target & $\sigma_{tgt}$ & 0.15 \\
 & Initial value & - & 0.01 \\
 \midrule
\multirow{5}{*}{\textbf{Optimizer}}  & Optimizer &  & Adam \\
 & Optimizer momentum & $(\beta_1, \beta_2)$ & (0.9, 0.999) \\
 & Learning rate scheduler & - & Cosine Decay \\
 & Learning rate init & $\eta$ & 3e-4 \\
 & Learning rate end & - & 1.5e-4 \\
 \midrule
\multirow{2}{*}{\textbf{Noise Repeat}} & Zeta distribution exponent & $s$ & 2 \\
 & Maximum repeat limit & - & 16 \\
    \bottomrule
\end{tabular}
}

%% file: tables/hyperparam_cpu.tex
\parbox{0.95\textwidth}{
\captionof{table}{\textbfs{Hyperparameters (CPU-based Simulators).} \textsc{FlashSAC} hyperparameters used in benchmarks with no parallel environments (MuJoCo, DMC, Humanoid Bench, and MyoSuite). We list only the values that differ from the settings provided in table~\ref{table:hyperparameters_gpusim}.}
\label{table:hyperparameters_cpusim}
\centering
\small
\begin{tabular}{llll}
\toprule
 & \textbf{Hyperparameter} & \textbf{Notation} & \textbf{Value} \\
 \midrule
\multirow{5}{*}{\textbf{Common}} 
 & Parallel environments & - & 1 \\
 & Replay buffer capacity & - & 1M \\
 & Batch size & - & 512 \\
 & Update-to-data (UTD) ratio & - & 1 \\
 & TD Steps (n-step) & $n$ & 1 \\
    \bottomrule
\end{tabular}
}

%% file: tables/hyperparam_visual.tex
\parbox{0.95\textwidth}{
\captionof{table}{\textbfs{Hyperparameters (Vision-Based RL).} \textsc{FlashSAC} hyperparameters used in vision-based tasks. We list only the values that differ from the settings provided in table~\ref{table:hyperparameters_gpusim}.}
\label{table:hyperparameters_visualrl}
\centering
\small
\begin{tabular}{llll}
    \toprule
     & \textbf{Hyperparameter} & \textbf{Notation} & \textbf{Value} \\
     \midrule
    \multirow{7}{*}{\textbf{Common}} 
     & Parallel environments & - & 1 \\
     & Replay buffer capacity & - & 1M \\
     & Batch size & - & 256 \\
     & Update-to-data (UTD) ratio & - & 0.5 \\
     & TD Steps (n-step) & $n$ & 3 \\
     & Action repeat & - & 2 \\
     & Frame stack & - & 3 \\
    \midrule
    \multirow{3}{*}{\textbf{Encoder}} & Number of layers & - & 4 \\
     & Number of channels & - & 32 \\
     & Output feature dim & - & 50 \\
    \bottomrule
    \end{tabular}
}

%% file: tables/info_sim2real_pdgain.tex
\parbox{0.95\textwidth}{
\captionof{table}{\textbfs{Joint Information of Unitree G1 humanoid.} Joint list of Unitree G1 29-DoF with default angle, stiffness $K_p$, and damping $K_d$, where we adopt heuristic parameters from~\cite{liao2025beyondmimic}.}
\label{tab:sim2real_pdgain}
\centering
\small
    \begin{tabular}{lccc}
        \toprule
        \textbf{Joint name} & \textbf{Default angle} & \textbf{$K_p$} & \textbf{$K_d$} \\
        \midrule
        \texttt{left\_hip\_pitch\_joint}   & $-0.2$ & $40.16$ & $2.559$ \\
        \texttt{left\_hip\_roll\_joint}    & $0$ & $99.08$ & $6.311$ \\
        \texttt{left\_hip\_yaw\_joint}     & $0$ & $40.16$ & $2.559$ \\
        \texttt{left\_knee\_joint}         & $0.42$ & $99.08$ & $6.311$ \\
        \texttt{left\_ankle\_pitch\_joint} & $-0.23$ & $28.49$ & $1.815$ \\
        \texttt{left\_ankle\_roll\_joint}  & $0$ & $28.49$ & $1.815$ \\
        \texttt{right\_hip\_pitch\_joint}     & $-0.2$ & $40.16$ & $2.559$ \\
        \texttt{right\_hip\_roll\_joint}      & $0$ & $99.08$ & $6.311$ \\
        \texttt{right\_hip\_yaw\_joint}       & $0$ & $40.16$ & $2.559$ \\
        \texttt{right\_knee\_joint}           & $0.42$ & $99.08$ & $6.311$ \\
        \texttt{right\_ankle\_pitch\_joint}   & $-0.23$ & $28.49$ & $1.815$ \\
        \texttt{right\_ankle\_roll\_joint}    & $0$ & $28.49$ & $1.815$ \\
        \texttt{waist\_yaw\_joint}            & $0$ & $40.16$ & $2.559$ \\
        \texttt{waist\_roll\_joint}            & $0$ & $28.49$ & $1.815$ \\
        \texttt{waist\_pitch\_joint}            & $0$ & $28.49$ & $1.815$ \\
        \texttt{left\_shoulder\_pitch\_joint} & $0.35$ & $14.25$ & $0.907$ \\
        \texttt{left\_shoulder\_roll\_joint}  & $0.18$ & $14.25$ & $0.907$ \\
        \texttt{left\_shoulder\_yaw\_joint}   & $0$ & $14.25$ & $0.907$ \\
        \texttt{left\_elbow\_joint}           & $0.87$ & $14.25$ & $0.907$ \\
        \texttt{left\_wrist\_roll\_joint}     & $0$ & $14.25$ & $0.907$ \\
        \texttt{left\_wrist\_pitch\_joint}    & $0$ & $16.78$ & $1.068$ \\
        \texttt{left\_wrist\_yaw\_joint}      & $0$ & $16.78$ & $1.068$ \\
        \texttt{right\_shoulder\_pitch\_joint}& $0.35$ & $14.25$ & $0.907$ \\
        \texttt{right\_shoulder\_roll\_joint} & $-0.18$ & $14.25$ & $0.907$ \\
        \texttt{right\_shoulder\_yaw\_joint}  & $0$ & $14.25$ & $0.907$ \\
        \texttt{right\_elbow\_joint}          & $0.87$ & $14.25$ & $0.907$ \\
        \texttt{right\_wrist\_roll\_joint}    & $0$ & $14.25$ & $0.907$ \\
        \texttt{right\_wrist\_pitch\_joint}   & $0$ & $16.78$ & $1.068$ \\
        \texttt{right\_wrist\_yaw\_joint}     & $0$ & $16.78$ & $1.068$ \\
        \bottomrule
    \end{tabular}
}

%% file: tables/info_sim2real_obs.tex
\parbox{0.95\textwidth}{
\captionof{table}{\textbfs{Observation Space for Sim-to-Real Experiments.} Our observation space combines proprioceptive
information about the robot state and the joint states. where base linear velocity $\boldsymbol{v}_t$ and exteroceptive height map $\boldsymbol{h}_t$ are privileged observations for critic. $\mathcal{U}$ indicates uniform distributions are used to augment the measurements and make the system robust against sensor noise.}
\label{tab:sim2real_obs}
\centering
\small
    \begin{tabular}{lccccc}
        \toprule
        \textbf{Observation}& \textbf{Notation} & \textbf{Dimension} & \textbf{Augmentation} & \textbf{Unit} \\
        \midrule
        Base linear velocity & $\boldsymbol{v}_t$ & $3$ & $\mathcal{U}[-0.1, 0.1]$ & m/s \\
        Base angular velocity & $\boldsymbol{\omega}_t$ & $3$ & $\mathcal{U}[-0.2, 0.2]$ & rad/s \\
        Projected gravity & $\boldsymbol{g}_t$ & $3$ & $\mathcal{U}[-0.05, 0.05]$ & $\text{m/s}^2$ \\
        Joint position & $\boldsymbol{q}_t$ & $29$ & $\mathcal{U}[-0.01, 0.01]$ & rad \\
        Joint velocity & $\boldsymbol{\dot{q}}_t$ & $29$ & $\mathcal{U}[-1.5, 1.5]$ & rad/s \\
        Last joint action & $\boldsymbol{a}_t$ & $29$ & - & rad \\
        Command velocity & $\boldsymbol{c}_t$ & $3$ & - & m/s, rad/s \\
        Height map & $\boldsymbol{h}_t$ & $17 \times 11$ & - & m \\
        \bottomrule
    \end{tabular}
}

%% file: tables/info_sim2real_reward.tex
\parbox{0.95\textwidth}{
\captionof{table}{\textbfs{Reward Configurations for Sim-to-Real Experiments.} Both methods share the same reward structure, including task rewards for tracking velocities, style rewards for its gait style and regularization terms penalizing excessive joint torques, action rate, and orientation instability. Different reward weights are applied to account for differing learning dynamics and ensure stable real-world deployment~\cite{seo2025fasttd3}.}
\label{tab:sim2real_reward}
\centering
\small
\begin{tabular}{llcc}
\toprule
\textbf{Reward} & \textbf{Expression} & \textbf{Weight (FlashSAC)} & \textbf{Weight (PPO)} \\
\midrule
\multicolumn{4}{l}{\textit{Task}} \\
Track linear velocity  &
$\exp \bigl(-\lVert \boldsymbol{v}_{xy}^{\mathrm{cmd}} - \boldsymbol{v}_{xy}^{\mathrm{yaw}}\rVert^2 / \sigma^2\bigr)$
& $2.0 \; (\sigma = 0.25)$ & $1.5 \; (\sigma = 0.5)$ \\
Track angular velocity  &
$\exp\bigl(-(\boldsymbol{\omega}_z^{\mathrm{cmd}} - \boldsymbol{\omega}_z)^2 / \sigma^2\bigr)$
& $1.5 \; (\sigma = 0.25)$ & $1.5 \; (\sigma = 0.5)$ \\
Orthogonal velocity  &
$\exp\bigl(-1.5\,\lVert \boldsymbol{v}_{\perp}\rVert^2\bigr)$
& $1.0$ & $1.0$ \\
\midrule
\multicolumn{4}{l}{\textit{Style}} \\
Feet air time      &
$r_{\mathrm{airtime}}(t_{\mathrm{swing}},\,t_{\mathrm{stance}},\,\lVert\boldsymbol{c}\rVert)$
& $1.0$ & $0.25$ \\
Feet slide         &
$\sum_{i=1}^2 \lVert \boldsymbol{v}_{xy}^{\mathrm{ft}_i}\rVert \cdot \mathbf{1}[c^{\mathrm{ft}_i}]$
& $-0.25$ & $-0.25$ \\
Feet yaw drag      &
$\sum_{i=1}^2 \lvert \boldsymbol{\omega}_z^{\mathrm{ft}_i}\rvert \cdot \mathbf{1}[c^{\mathrm{ft}_i}]$
& $-0.5$ & $-0.25$ \\
Feet force         &
$\mathrm{clamp}\bigl(\lVert \boldsymbol{f}_z^{\mathrm{ft}}\rVert - 700,\;0,\;400\bigr)$
& $-3{\times}10^{-3}$ & $-3{\times}10^{-3}$ \\
Feet lateral distance      &
$\mathrm{clamp}\bigl(0.3 - d_{\mathrm{lateral}},\;0,\;\infty\bigr)$
& $-5.0$ & $-2.0$ \\
Feet stumble       &
$\mathbf{1}\bigl[\lVert \boldsymbol{f}_{xy}^{\mathrm{ft}}\rVert > 5\lvert \boldsymbol{f}_z^{\mathrm{ft}}\rvert\bigr]$
& $-2.0$ & $-2.0$ \\
Air time variance  &
$\mathrm{Var}[t_{\mathrm{air}}] + \mathrm{Var}[t_{\mathrm{contact}}]$
& $-1.0$ & $-1.0$ \\
Impact velocity delta  &
$\sum_{i=1}^2 \min\bigl(\Delta \boldsymbol{v}_{z, \mathrm{ft}_i}^{2},\;1\bigr)$
& $-5.0$ & $-5.0$ \\
\midrule
\multicolumn{4}{l}{\textit{Regularization}} \\
Linear velocity ($z$)  & $\boldsymbol{v}_z^2$ & $-0.25$ & $-0.25$ \\
Angular velocity ($xy$)& $\lVert \boldsymbol{\omega}_{xy}\rVert^2$ & $-1.0$ & $-0.05$ \\
Energy             & $\lVert \boldsymbol{\tau} \odot \dot{\boldsymbol{q}}\rVert$
& $-10^{-3}$ & $-10^{-3}$ \\
Joint acceleration         & $\lVert \ddot{\boldsymbol{q}}\rVert^2$
& $-2.5{\times}10^{-7}$ & $-2.5{\times}10^{-7}$ \\
Action rate        & $\lVert \boldsymbol{a}_t - \boldsymbol{a}_{t-1}\rVert^2$
& $-0.5$ & $-0.01$ \\
Flat orientation   & $\lVert \boldsymbol{g}_{xy}^{\mathrm{base}}\rVert^2$
& $-5.0$ & $-1.0$ \\
Body orientation   & $\lVert \boldsymbol{g}_{xy}^{\mathrm{torso}}\rVert^2$
& $-52.0$ & $-2.0$ \\
Joint deviation   & $\lVert \boldsymbol{q} - \boldsymbol{q}_0\rVert_1$
& \makecell[t]{
$-0.25$ (leg) \\
$-0.50$ (hip) \\
$-1.00$ (arm)
} & \makecell[t]{
$-0.02$ (leg) \\
$-0.15$ (hip) \\
$-0.20$ (arm)
} \\
Joint position limits  & $\max(\boldsymbol{q}-\boldsymbol{q}_{\text{max}},0)
+\max(\boldsymbol{q}_{\text{min}}-\boldsymbol{q},0)$
& $-5.0$ & $-2.0$ \\
Stand still        &
$\lVert \boldsymbol{q} - \boldsymbol{q}_0\rVert^2
\cdot \mathbf{1}\bigl[\lVert\boldsymbol{c}\rVert < 0.1\bigr]$
& $-5.0$ & $-0.25$ \\
\midrule
\multicolumn{4}{l}{\textit{Safety \& Termination}} \\
Undesired contacts &
$\sum_i \mathbf{1}\bigl[\lVert \boldsymbol{f}_i\rVert > 1\bigr]$ (non-ankle)
& $-5.0$ & $-1.0$ \\
Fly                &
$\mathbf{1}[\texttt{no ankle contact}]$
& $-1.0$ & $-1.0$ \\
Termination penalty &
$\mathbf{1}[\texttt{terminated} \wedge \neg\;\texttt{timeout}]$
& - & $-200$ \\
Alive bonus        &
$\mathbf{1}[\neg \; \texttt{terminated}]$
& $1.0$ & - \\
\bottomrule
\end{tabular}
}

%% file: tables/info_sim2real_reward_symbol.tex
\parbox{0.95\textwidth}{
\captionof{table}{\textbfs{Notation for Reward Terms.} Symbol definitions of the reward terms in Table~\ref{tab:sim2real_reward}.}
\label{tab:sim2real_reward_symbol}
    \small
    \centering
    \begin{tabular}{lp{12cm}}
    \toprule[1.0pt]
    \textbf{Symbol} & \textbf{Description} \\
    \midrule[0.8pt]
    $\boldsymbol{c} = [\boldsymbol{v}_{xy}^{\mathrm{cmd}} \; \boldsymbol{\omega}_z^{\mathrm{cmd}}]$ & Commanded $xy$ linear velocity and yaw angular velocity from the command velocity $\boldsymbol{c}$. \\ [0.2ex]
    $\boldsymbol{v}_{xy}^{\mathrm{yaw}}, \boldsymbol{\omega}_z$ & Robot $xy$ linear velocity projected into the yaw-aligned body frame and yaw angular velocity. \\ [0.2ex]
    $\boldsymbol{v}_{\perp}$ & \makecell[tl]{Velocity component orthogonal to the commanded direction; \\ for stop commands ($\lVert\boldsymbol{c}\rVert < 0.1$), equals $\boldsymbol{v}_{xy}^{\mathrm{yaw}}$} \\ [0.2ex]
    $\boldsymbol{q}_0$ & Default joint positions. \\ [0.2ex]
    $(\boldsymbol{q}_{\min}, \boldsymbol{q}_{\max})$ & Soft joint position limit for joint $j$. \\ [0.2ex]
    $\boldsymbol{g}_{xy}^{\mathrm{base}}$, $\boldsymbol{g}_{xy}^{\mathrm{torso}}$ & \makecell[tl]{$xy$ components of the gravity vector projected into the base and torso body frames; \\ $\mathbf{0}$ when perfectly upright.} \\ [0.2ex] 
    $\boldsymbol{f}_i$ & Net contact force on body $i$ in world frame. \\ [0.2ex]
    $\boldsymbol{f}_{xy}^{\mathrm{ft}}$, $\boldsymbol{f}_z^{\mathrm{ft}}$ & Horizontal and vertical components of foot contact force, respectively. \\ [0.2ex]
    $c^{\mathrm{ft}_i}$ & Binary contact state of foot $i$, true when $\lVert\boldsymbol{f}^{\mathrm{ft}_i} \rVert > 1\mathrm{N}$. \\ [0.2ex]
    $\boldsymbol{v}_{xy}^{\mathrm{ft}_i}, \boldsymbol{\omega}_z^{\mathrm{ft}_i}$ & $xy$ linear and yaw angular velocity of foot $i$ in world frame, respectively \\ [0.2ex]
    $d_{\mathrm{lateral}}$ & Lateral ($y$-axis) distance between left and right feet in the body frame; threshold set to $0.3\;\mathrm{m}$. \\ [0.2ex]
    $\Delta \boldsymbol{v}_{z,\mathrm{ft}_i}$ & Frame-to-frame change in vertical velocity of foot $i$; clamped at $\Delta v_{\max}{=}1.0\;\mathrm{m/s}$. \\ [0.2ex]
    $t_{\mathrm{swing}}$, $t_{\mathrm{stance}}$ & Current air (swing) time and ground contact (stance) time per foot. \\ [0.2ex]
    $t_{\mathrm{air}}$, $t_{\mathrm{contact}}$ & Last completed air and contact durations per foot, used for variance computation. \\ [0.2ex]
    $r_{\mathrm{airtime}}$ & \makecell[tl]{Composite feet air-time reward with in-place handling; blends swing-target ($0.4\;\mathrm{sec}$) \\, stance-target ($0.5 / \lVert\boldsymbol{c}\rVert$, clipped to $[0.1,\,0.5]\;\mathrm{sec}$), and in-place mode ($\lVert\boldsymbol{c}\rVert < 0.25$).} \\ [0.2ex]
    $\mathbf{1}[\cdot]$ & Indicator function, returning $1$ when the condition is true and $0$ otherwise. \\ [0.2ex]
    \bottomrule[1.0pt]
    \end{tabular}
}